\documentclass[10pt,twocolumn,letterpaper]{article}

\usepackage[pagenumbers]{iccv} 

\definecolor{iccvblue}{rgb}{0.21,0.49,0.74}
\usepackage[pagebackref,breaklinks,colorlinks,allcolors=iccvblue]{hyperref}
\usepackage{stfloats} 
\newcommand{\redref}[1]{%
  {\hypersetup{linkcolor=red}%
   \ref{#1}}%
  \hypersetup{linkcolor=blue}
}

\def\paperID{*****} 
\def\confName{ICCV}
\def\confYear{2025}

\title{BioDet: Boosting Industrial Object Detection with Image Preprocessing Strategies}

\author{
Jiaqi Hu$^{1}$ \quad
Hongli Xu$^{1}$ \quad
Junwen Huang$^{1}$ \quad
Peter KT Yu$^{2}$\\
Slobodan Ilic$^{1}$ \quad
Benjamin Busam$^{1}$ \\
\\
$^{1}$Technical University of Munich \quad
$^{2}$XYZ Robotics \\
{\tt\small \{jiaqi.hu, hongli.xu, junwen.huang, slobodan.ilic, b.busam\}@tum.de},\\
{\tt\small peter.yu@xyzrobotics.com}\\
{\small \href{https://github.com/jacky-hjqq/BioDet}{\texttt{\textbf{\textcolor{magenta}{github.com/jacky-hjqq/BioDet}}}}}
}

\usepackage{makecell}
\usepackage{hyperref}
\usepackage{xcolor}

\begin{document}
\setlength{\parskip}{0em} 
\maketitle
\begin{abstract}
Accurate 6D pose estimation is essential for robotic manipulation in industrial environments. Existing pipelines typically rely on off-the-shelf object detectors followed by cropping and pose refinement, but their performance degrades under challenging conditions such as clutter, poor lighting, and complex backgrounds—making detection the critical bottleneck. In this work, we introduce a standardized and plug-in pipeline for 2D detection of unseen objects in industrial settings. Based on the current SOTA baselines, our approach reduces domain shift and background artifacts through (i) low-light image enhancement, (ii) background removal guided by open-vocabulary detection with foundation models. This design suppresses the false positives prevalent in raw SAM outputs, yielding more reliable detections for downstream pose estimation. Extensive experiments on real-world industrial bin-picking benchmarks from BOP demonstrate that our method significantly boosts the detection accuracy while incurring negligible inference overhead, showing the effectiveness and practicality of the proposed method.
\end{abstract}    
\section{Introduction}
\label{sec:intro}

\begin{figure*}[htbp]
  \centering
  \includegraphics[width=\textwidth]{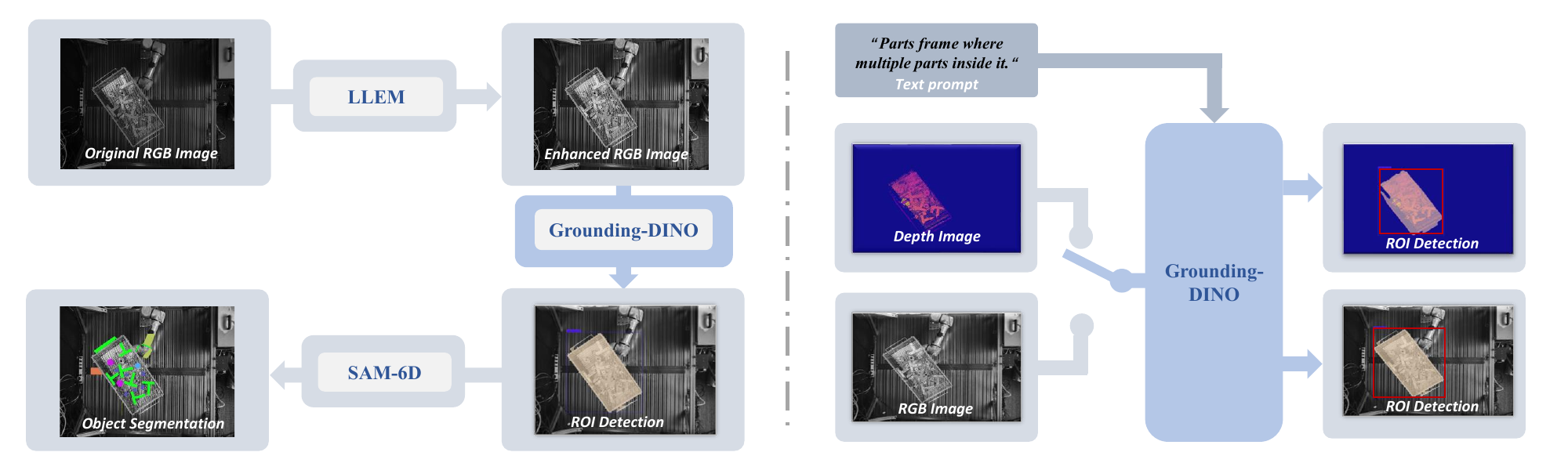}
    \caption{(Left) Overview of our pipeline. Images from industrial datasets~\cite{Kalra_2024_CVPR,huang2025xyz,Drost_2017_ICCV} are first enhanced by our Low-Light Enhancement Module (LLEM)~\cite{yan2025hvi}, then processed by Grounding-DINO~\cite{liu2023grounding} for ROI detection; the highest-confidence crop is passed to SAM6D~\cite{lin2023sam6d} for final segmentation. (Right) ROI detection example. Either a plasma-mapped depth image or the raw RGB input is prompted with “Parts frame where multiple parts are inside it” to generate a bounding box for background removal.}

  \label{fig:pipeline}
\end{figure*}

Research on 6D pose estimation has largely focused on everyday household objects, as reflected in the benchmarks of BOP~\cite{sundermeyer2023bopchallenge, nguyen2025bop}. While state-of-the-art methods achieve strong accuracy under these controlled conditions, their performance drops significantly in industrial environments characterized by extreme lighting, heavy clutter, and severe object stacking~\cite{huang2025xyz,Kalra_2024_CVPR,Drost_2017_ICCV}. In practice, most robotic manipulation pipelines in factories still rely on overfitting networks to specific target objects in order to achieve the required precision. Thus, enabling robots with generalizable and reliable 6DoF pose estimation across diverse industrial setups remains a critical yet unresolved challenge for flexible automation. To highlight these issues, new industrial benchmarks such as XYZ-IBD~\cite{huang2025xyz} and IPD~\cite{Kalra_2024_CVPR} have been introduced. XYZ-IBD emphasizes reflective surfaces, symmetric shapes, dense occlusions, and cluttered stacking, while IPD focuses on extreme HDR lighting. These datasets have renewed attention on the unique challenges of industrial 6D pose estimation and motivated research into robustness under such demanding conditions.

Current generalized (unseen-object) model-based pipelines~\cite{wen2024foundationpose,lin2023sam6d,huang2024matchu,nguyen2024gigapose} typically assume access to a CAD model at test time and follow a two-stage process: (1) detecting and localizing the object in the input image, and (2) estimating its 6D pose from the cropped region. Prior work has mainly emphasized the second stage, while relying on predefined detection or segmentation for the first. As a result, pose estimation accuracy heavily depends on detection quality—failures in object detection directly propagate to pose errors, an issue that is particularly problematic in industrial scenes with severe clutter and object ambiguities.

Recent advances in foundation models, such as SAM~\cite{kirillov2023segany} and DINOv2~\cite{oquab2023dinov2}, have transformed the detection stage in unseen-object pipelines. Approaches like CNOS~\cite{nguyen2023cnos} and SAM-6D~\cite{lin2023sam6d} leverage these models to localize arbitrary objects without task-specific training, making them the de facto choice for detection. These methods render CAD templates and compute feature similarity with SAM-generated segments. However, their success is highly dependent on segmentation quality and template matching accuracy. Both are prone to failure under low contrast edges, poor lighting, reflective surfaces, or ambiguous geometries, \ie conditions that are ubiquitous in industrial environments. In practice, SAM generates thousands of candidate segments per image, ranked by similarity to template features. Detection failures may arise from incorrect template alignment, under- or over-segmentation, or mismatched reflective regions. Interestingly, we observe that once the object identity is correct, even imperfect masks often still yield reliable pose estimation~\cite{lin2023sam6d, wen2024foundationpose}, underscoring the importance of robust template-to-segment matching. For industrial applications, this matching step must overcome severe domain gaps between rendered templates and real-world images captured under challenging lighting and clutter. 

We address this challenge from two perspectives: reducing the domain gap between image segments and rendered templates, and suppressing redundant background segments to minimize false candidates for object matching.
We propose a robust pipeline for model-based unseen object detection in industrial scenarios. First, we deploy an image enhancement network to mitigate adverse lighting conditions (e.g., under/over-exposure, strong reflections). Enhanced images improve both SAM’s segmentation quality and the scoring module’s ability to distinguish target objects. Second, we employ the open-vocabulary detector Grounding-DINO~\cite{liu2023grounding} with task-adaptive prompts (e.g., “the object frame fixed by a robot arm”) to filter background regions and eliminate invalid mask proposals. This reduces the risk of background clutter being incorrectly matched to templates. Finally, we perform template matching with DINOv2 features following~\cite{nguyen2023cnos, lin2023sam6d}. Our pipeline substantially improves object detection robustness and, consequently, pose estimation accuracy. On industrial datasets, it significantly outperforms baseline models that do not incorporate image enhancement or contextual filtering.
\section{Related Work}
\label{sec:related work}

\subsection{Low-Light Image Enhancement}
Low-Light Image Enhancement (LLIE) aims to recover visual details from underexposed images. Most LLIE methods operate in RGB space, but its sensitivity to hue often causes color shifts and brightness artifacts. Transforming to HSV space mitigates brightness issues, yet typically introduces red or black noise. Recent work, HVI~\cite{yan2025hvi}, addresses these limitations by introducing a novel color space that combines polarized hue and saturation with a learnable intensity channel, and proposes CIDNet to achieve state-of-the-art results. LLIE is increasingly applied in downstream domains: for example, LighTDiff~\cite{chen2024lightdiff} improves surgical frames prior to segmentation or diagnosis, while in autonomous driving, LLIE is used to enhance fisheye-camera inputs for object detection in traffic monitoring~\cite{tran2024low}.

\subsection{Segmentation of Unseen Objects}
The Segment Anything Model (SAM)~\cite{kirillov2023segany}, trained on the large-scale SA-1B dataset, is a powerful foundation model that supports flexible prompts (points, boxes, or text) and generalizes in a zero-shot manner across diverse domains, including medical imaging~\cite{ma2023medsam, mazurowski2023sammedical, zhang2023improvinggeneralization}, camouflaged object detection~\cite{ji2023samconcealed, tang2023cansamcod}, and transparent object segmentation~\cite{han2023samglass, ji2024samapplications}. Despite producing high-quality masks, SAM’s large model size leads to slow inference. FastSAM~\cite{zhao2023fastsam} alleviates this by employing a convolutional backbone trained on only 1/50 of SA-1B with far fewer parameters, achieving comparable segmentation accuracy at nearly 50× speed. Both SAM and FastSAM generate high-quality masks useful for downstream tasks like 6D pose estimation. However, their raw outputs often include noisy segments, increasing complexity for template matching. Grounding-DINO~\cite{liu2023grounding} partially mitigates this by conditioning SAM outputs with text prompts describing object categories or attributes.

Current model-based pipelines for unseen object segmentation~\cite{nguyen2023cnos,lin2023sam6d,chen2023zeropose} typically adopt SAM or FastSAM to generate proposals by randomly sampling points in the input image, then use DINOv2 features to match proposals with rendered templates. ZeroPose~\cite{chen2023zeropose} combines SAM proposals with visual and geometric embeddings to match each mask to its closest template and assign a CAD ID. CNOS~\cite{nguyen2023cnos} instead uses the DINOv2 class token and selects the CAD model ID with the highest average similarity. SAM6D~\cite{lin2023sam6d} introduces a multi-level similarity metric that integrates semantic, appearance, and geometric cues for object-template alignment. While effective in controlled household settings, these approaches degrade under industrial conditions such as those in XYZ-IBD~\cite{huang2025xyz} and IPD~\cite{Kalra_2024_CVPR}, where lighting, clutter, and distributional shifts pose greater challenges. Specifically, the failure of the object matching stage leads to the incorrect object identification in the scene, which will further cause the failure of the follow-up pose estimation stage. 


\section{Method}

In this section we present our image preprocessing strategies for industrial object detection. Section \redref{sec:llem} describes our low-light enhancement model, which restores images captured under extreme lighting. Section \redref{sec:background remove} details our background removal model, which isolates the target object from chaotic surroundings. Finally, Section \redref{sec:object detection} covers the matching stage, in which object masks are retrieved and assigned CAD-model labels based on their matching scores.

\subsection{Low Light Enhancement}
\label{sec:llem}
In factory bin-picking scenarios, industrial cameras often capture high-quality grayscale images. This creates a domain gap for foundational models such as SAM and DINO, which are primarily trained on large-scale RGB datasets. Moreover, metallic parts introduce textureless surfaces, strong reflections, and severe multi-instance clutter, further complicating recognition and detection. To address these challenges, we enhance the input images prior to applying pretrained models, improving boundary visibility under challenging lighting.

Specifically, we first evaluate the brightness of each query image by averaging pixel intensities. If the measured intensity falls below a predefined threshold, a low-light enhancement model is applied; otherwise, the image remains unchanged. For enhancement, we adopt the HVI model~\cite{yan2025hvi}, which introduces a novel color space and a Color and Intensity Decoupling Network (CIDNet)~\cite{yan2025hvi} to overcome color bias and brightness artifacts. CIDNet leverages a learnable intensity channel with polarized hue and saturation maps to restore details in underexposed regions. This selective strategy avoids unnecessary processing of well-lit frames and prevents overexposure of reflective metallic parts, yielding more reliable inputs for downstream segmentation and pose estimation.

\subsection{Background Removal}
\label{sec:background remove}
Raw SAM predictions yield thousands of segments across the entire image, introducing substantial computational overhead for template matching and increasing the likelihood of false positives due to artifacts.

To mitigate this, we incorporate Grounding-DINO~\cite{liu2023grounding} for background removal on the enhanced RGB images. In the industrial bin-picking setup, objects are typically centralized in a container or platform. We therefore issue the prompt “Parts frame where multiple parts inside it”, which reliably produces a single bounding box covering the region of interest (ROI) and suppresses surrounding clutter.

The predicted bounding box for the ROI localization is applied to crop both the RGB and depth inputs. To maintain geometric consistency, we adjust the camera intrinsics to compensate for the shifted image origin. The bounding box is stored, enabling accurate remapping of the detection and pose estimation results back to the original image coordinates.

\subsection{Object Detection}
\label{sec:object detection}

After extracting the ROI, we apply the Segment Anything Model (SAM)~\cite{kirillov2023segany} to generate a diverse set of candidate masks. To ensure full coverage, we sample a uniform grid of points across the cropped image and use them as prompts. The resulting proposals are filtered by discarding those below a confidence threshold and applying non-maximum suppression to remove redundant or highly overlapping masks. This process yields a compact set of high-quality candidate segments.

For each mask proposal, we assign an object ID by computing a matching score against rendered object templates. We render multiple templates by sampling object poses in SE(3) and extract both global class tokens and local patch features using a pre-trained DINOv2 ViT~\cite{oquab2023dinov2}. 

Following the SAM6D framework~\cite{lin2023sam6d} each mask proposal is similarly encoded to produce comparable embeddings. Specifically, we evaluate the similarity between proposals and templates along three complementary dimensions. Semantic similarity measures the alignment of global embeddings and provides a coarse object identity. Appearance similarity compares patch-level features, enabling fine-grained discrimination between visually similar objects. Geometric consistency estimates whether the spatial alignment of the rendered template is consistent with the location and extent of the proposal, which is particularly useful for resolving ambiguities under clutter and occlusion.

Finally, these three scores are integrated into a unified matching metric, where higher weights are placed on reliable cues such as geometry under visible regions. The template with the highest score is assigned as the final object label for the proposal, and its associated 6D pose hypothesis is retained for downstream refinement.

\section{Experiments}
\begin{table*}[!t]
  \centering
  \begin{tabular}{l|l|l|l|l|l}
    \toprule
    Methods & Segmentation Model & \multicolumn{4}{c}{Dataset} \\
    \cmidrule(lr){3-6}
    & & IPD & XYZ-IBD & ITODD & Mean \\
    \midrule
    CNOS                & FastSAM & 20.8 & 21.4 & 32.5 (BOP) & 24.9 \\
    CNOS                & SAM     & 15.6 & 17.1 & 31.3 (BOP) & 21.3 \\
    SAM6D               & FastSAM & 28.6 & 23.1 & 41.9 (BOP) & 31.2 \\
    SAM6D               & SAM     & 30.0 & 25.7 & 39.4 (BOP) & 31.7 \\
    Ours (w/o G-DINO)   & FastSAM & 28.8 (+0.2) & 23.2 (+0.1) & — & \\
    Ours (w/o LLEM)     & FastSAM & 40.9 (+12.3) & 25.4 (+2.3) & — & \\
    Ours                & FastSAM & 42.2 \textcolor{green}{(+13.6)} & 26.0 \textcolor{green}{(+2.9)} & 41.9 (BOP) & 36.7 \textcolor{green}{(+5.5)} \\
    Ours                & SAM     & 47.3 \textcolor{green}{(+17.3)} & 27.0 \textcolor{green}{(+1.3)} & 39.4 (BOP) & 37.9 \textcolor{green}{(+6.2)} \\
    \bottomrule
  \end{tabular}
  \caption{Quantitative comparison on the IPD \cite{Kalra_2024_CVPR}, XYZ-IBD \cite{huang2025xyz}, and ITODD \cite{Drost_2017_ICCV} datasets for CNOS \cite{nguyen2023cnos} and SAM6D \cite{lin2023sam6d}. We report the AP metric (higher is better) using the protocol from \cite{sundermeyer2023bopchallenge}. With our image preprocessing strategies, we outperform other methods. We also conduct ablation experiments with FastSAM to evaluate the efficacy of each individual preprocessing strategy. The ITODD dataset features good lighting and no background clutter; therefore, we use the baseline model without any preprocessing and the results are from BOP challenge \cite{sundermeyer2023bopchallenge} leaderboards.} \vspace{0.5cm}
  \label{tab:results}
\end{table*}

\begin{figure*}[t]
  \centering
  \begin{subfigure}[t]{0.24\linewidth}
    \includegraphics[width=\linewidth]{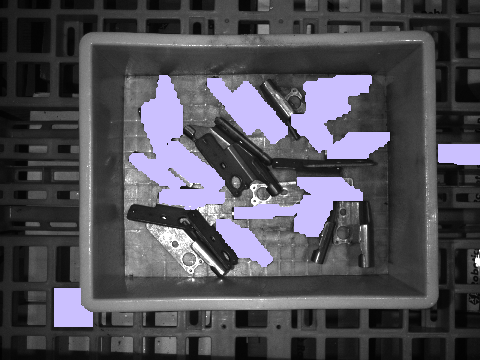}
  \end{subfigure}\hfill
  \begin{subfigure}[t]{0.24\linewidth}
    \includegraphics[width=\linewidth]{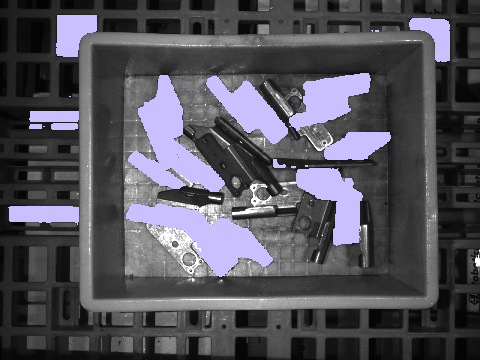}
  \end{subfigure}\hfill
  \begin{subfigure}[t]{0.24\linewidth}
    \includegraphics[width=\linewidth]{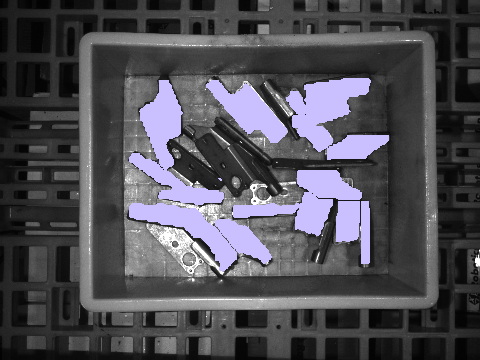}
  \end{subfigure}\hfill
  \begin{subfigure}[t]{0.24\linewidth}
    \includegraphics[width=\linewidth]{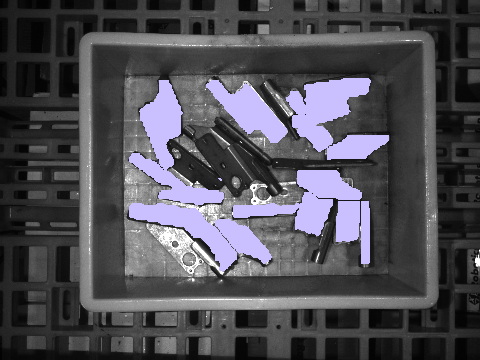}
  \end{subfigure}

  \vspace{1.5mm}

  \begin{subfigure}[t]{0.24\linewidth}
    \includegraphics[width=\linewidth]{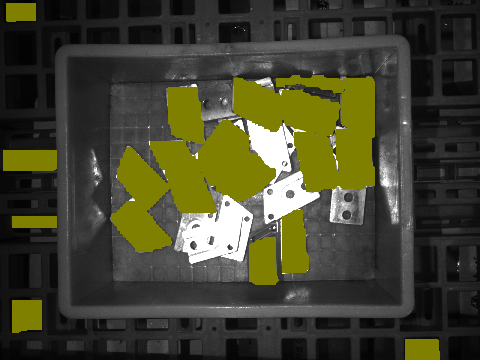}
  \end{subfigure}\hfill
  \begin{subfigure}[t]{0.24\linewidth}
    \includegraphics[width=\linewidth]{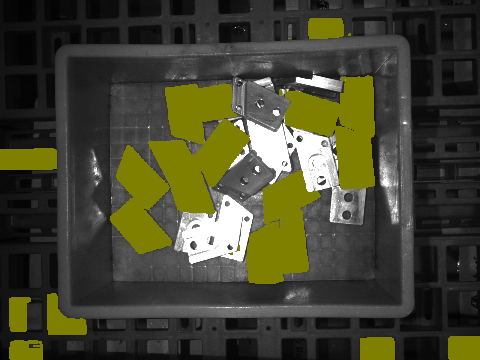}
  \end{subfigure}\hfill
  \begin{subfigure}[t]{0.24\linewidth}
    \includegraphics[width=\linewidth]{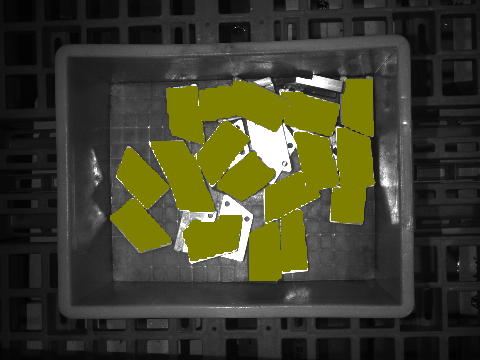}
  \end{subfigure}\hfill
  \begin{subfigure}[t]{0.24\linewidth}
    \includegraphics[width=\linewidth]{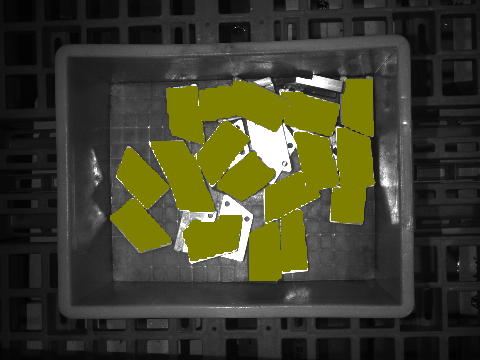}
  \end{subfigure}

  \vspace{1.5mm}

  \begin{subfigure}[t]{0.24\linewidth}
    \includegraphics[width=\linewidth]{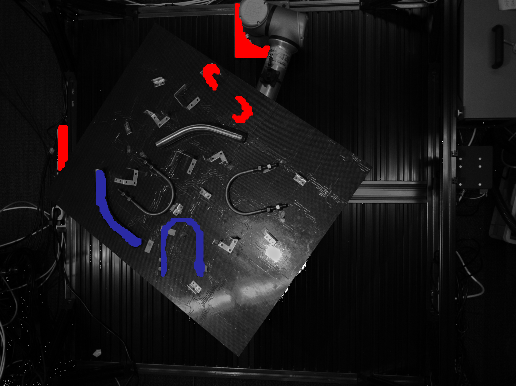}
  \end{subfigure}\hfill
  \begin{subfigure}[t]{0.24\linewidth}
    \includegraphics[width=\linewidth]{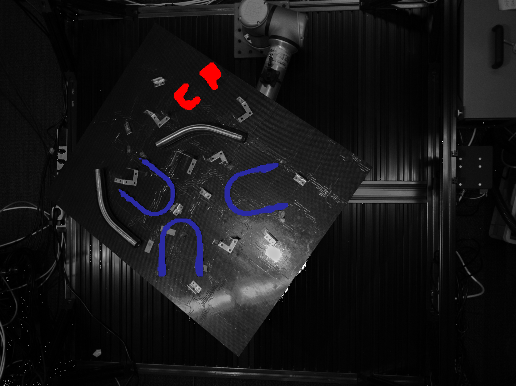}
  \end{subfigure}\hfill
  \begin{subfigure}[t]{0.24\linewidth}
    \includegraphics[width=\linewidth]{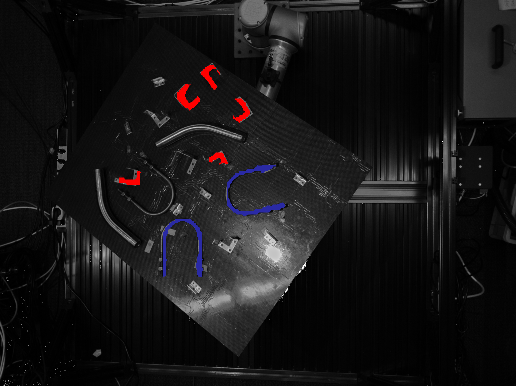}
  \end{subfigure}\hfill
  \begin{subfigure}[t]{0.24\linewidth}
    \includegraphics[width=\linewidth]{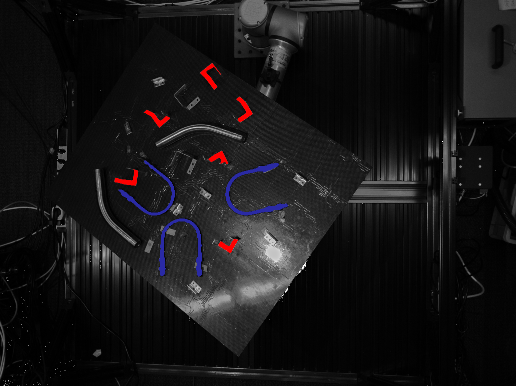}
  \end{subfigure}

  \vspace{1.5mm}

  \begin{subfigure}[t]{0.24\linewidth}
    \includegraphics[width=\linewidth]{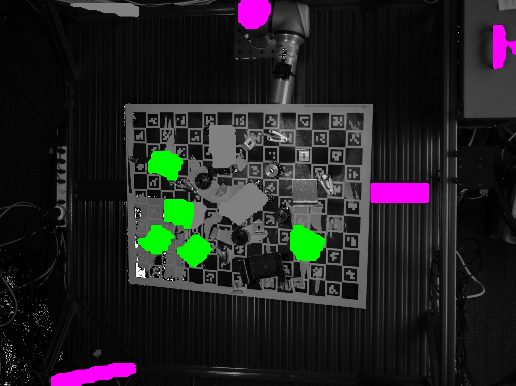}
    \caption{SAM6D (FastSAM)}
    \label{fig:ipd:col1}
  \end{subfigure}\hfill
  \begin{subfigure}[t]{0.24\linewidth}
    \includegraphics[width=\linewidth]{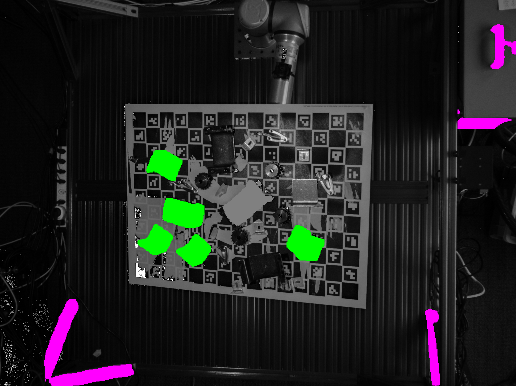}
    \caption{SAM6D (SAM)}
    \label{fig:ipd:col2}
  \end{subfigure}\hfill
  \begin{subfigure}[t]{0.24\linewidth}
    \includegraphics[width=\linewidth]{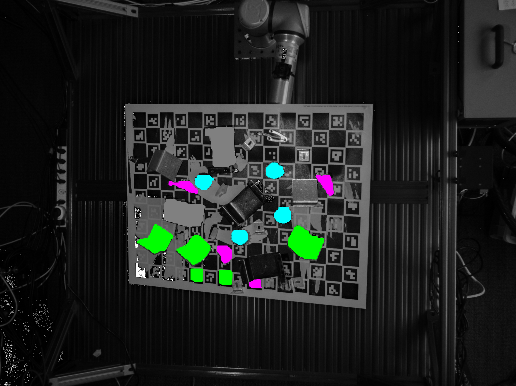}
    \caption{Ours (FastSAM)}
    \label{fig:ipd:col3}
  \end{subfigure}\hfill
  \begin{subfigure}[t]{0.24\linewidth}
    \includegraphics[width=\linewidth]{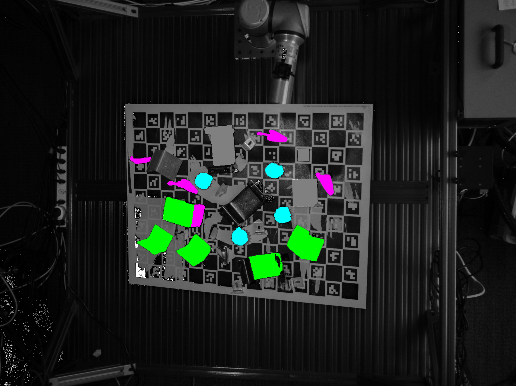}
    \caption{Ours (SAM)}
    \label{fig:ipd:col4}
  \end{subfigure}

  \caption{Qualitative results on the XYZ–IBD \cite{huang2025xyz} (top row) and IPD \cite{Kalra_2024_CVPR} (bottom row) datasets. Compared to the SOTA baseline method SAM6D~\cite{lin2023sam6d}, our detection pipeline exhibits fewer false positive detections and more true positives, demonstrating the effectiveness of our image enhancement and background removal modules under such industrial bin-picking setups. }
  \label{fig:combined_results}
\end{figure*}

\begin{figure}[t]
  \centering
  \begin{subfigure}[b]{0.32\linewidth}
    \includegraphics[width=\linewidth]{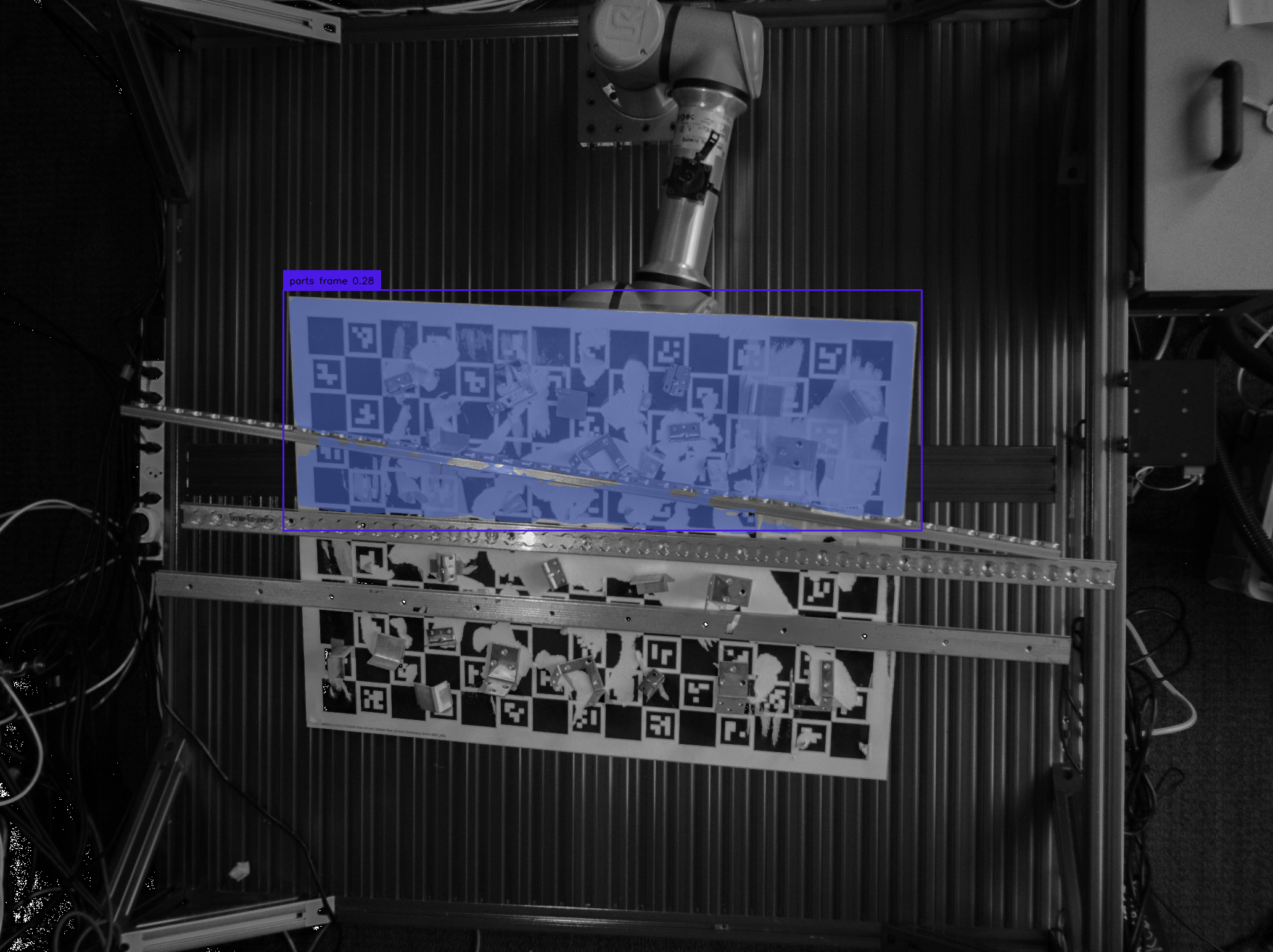}
  \end{subfigure}
  \begin{subfigure}[b]{0.32\linewidth}
    \includegraphics[width=\linewidth]{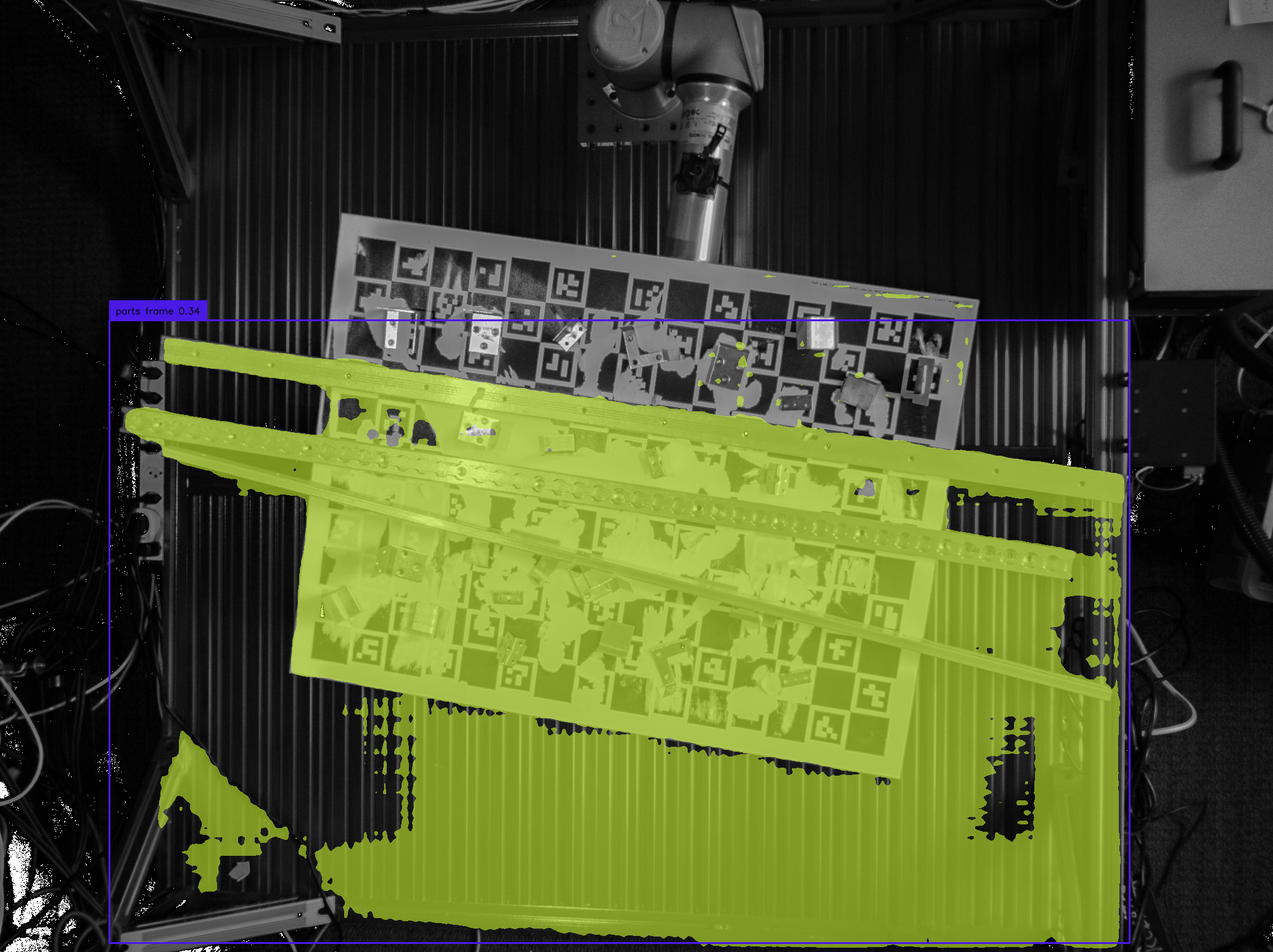}
  \end{subfigure}
  \begin{subfigure}[b]{0.32\linewidth}
    \includegraphics[width=\linewidth]{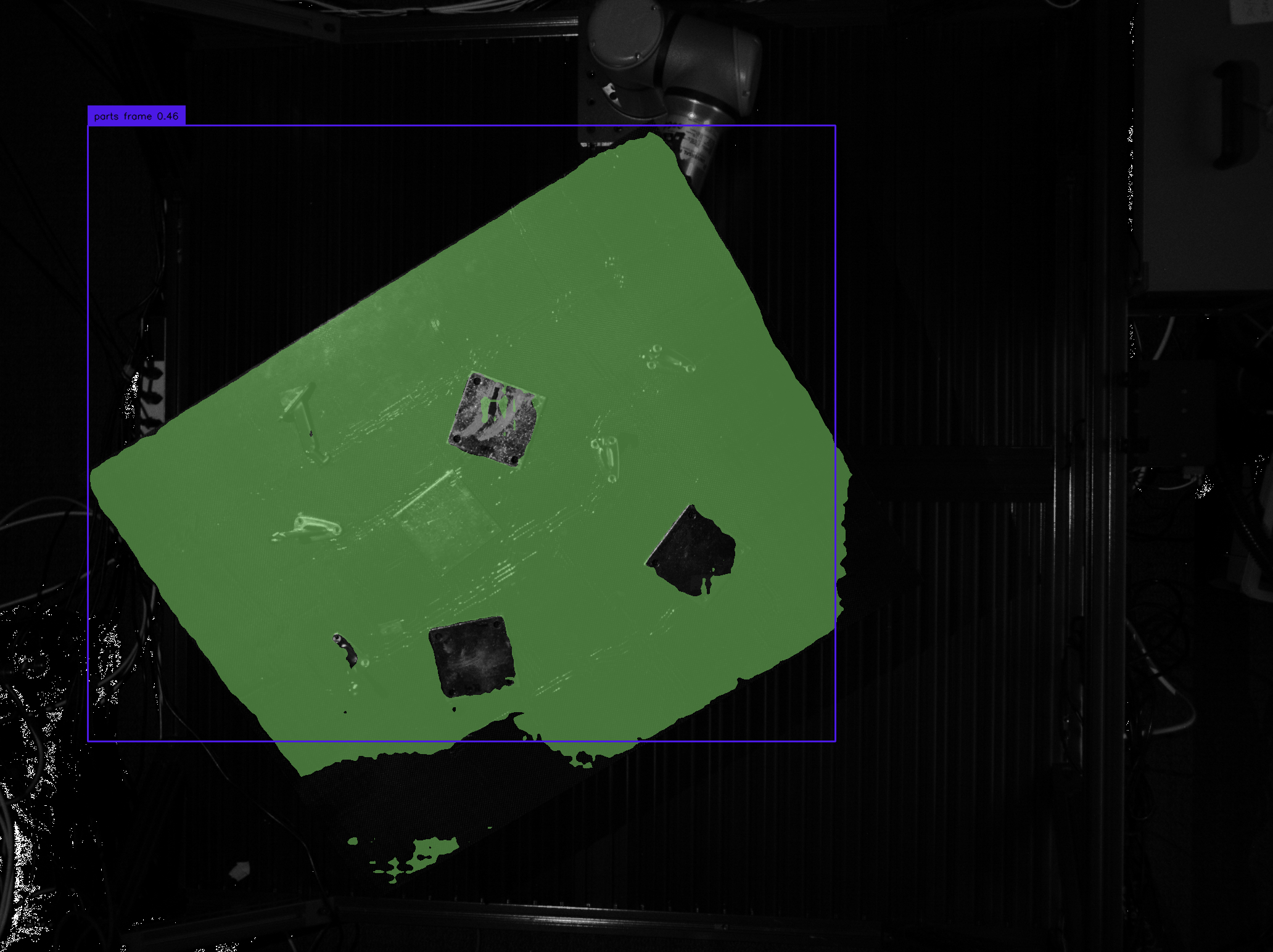}
  \end{subfigure}

  \begin{subfigure}[b]{0.32\linewidth}
    \includegraphics[width=\linewidth]{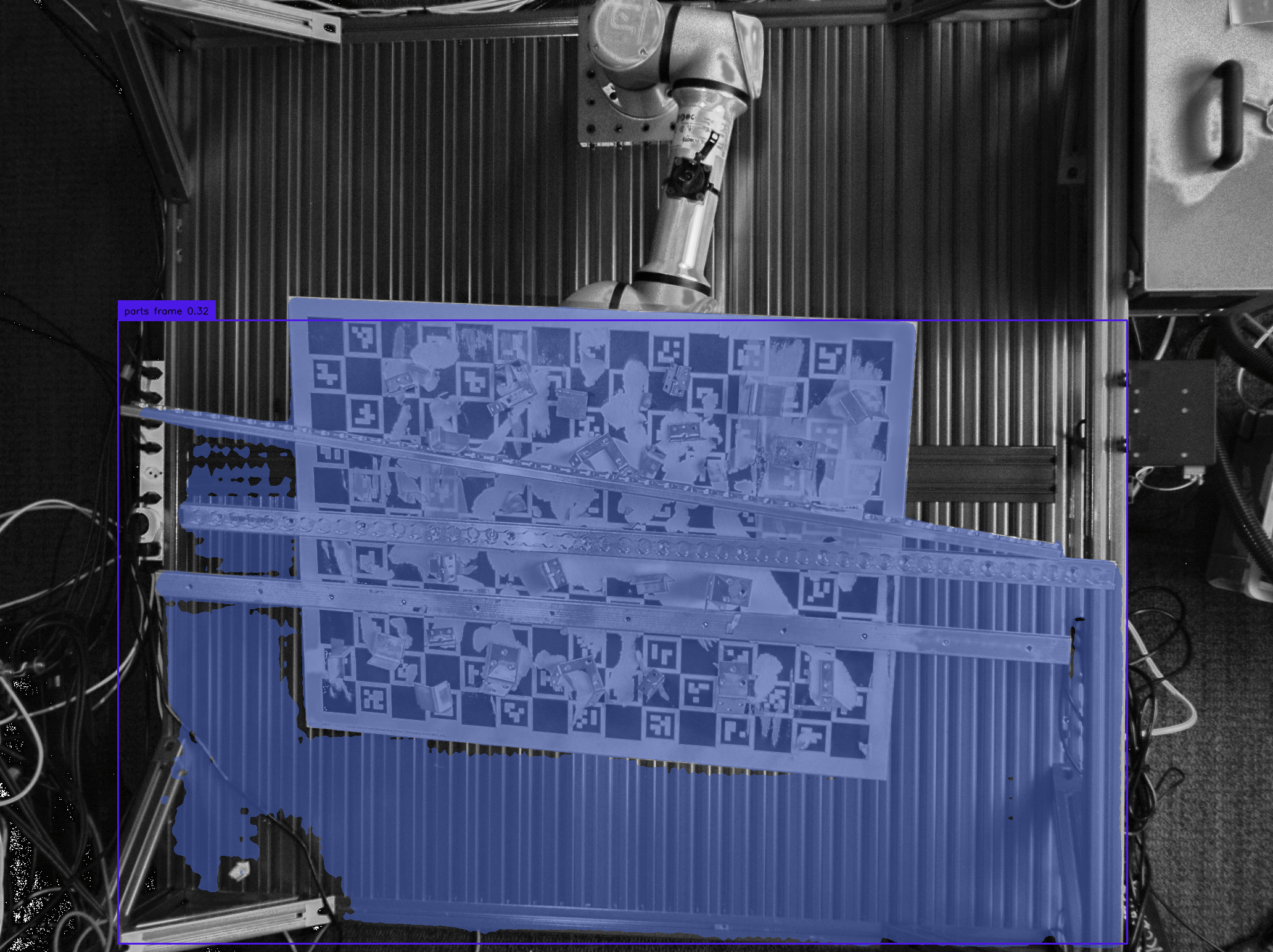}
  \end{subfigure}
  \begin{subfigure}[b]{0.32\linewidth}
    \includegraphics[width=\linewidth]{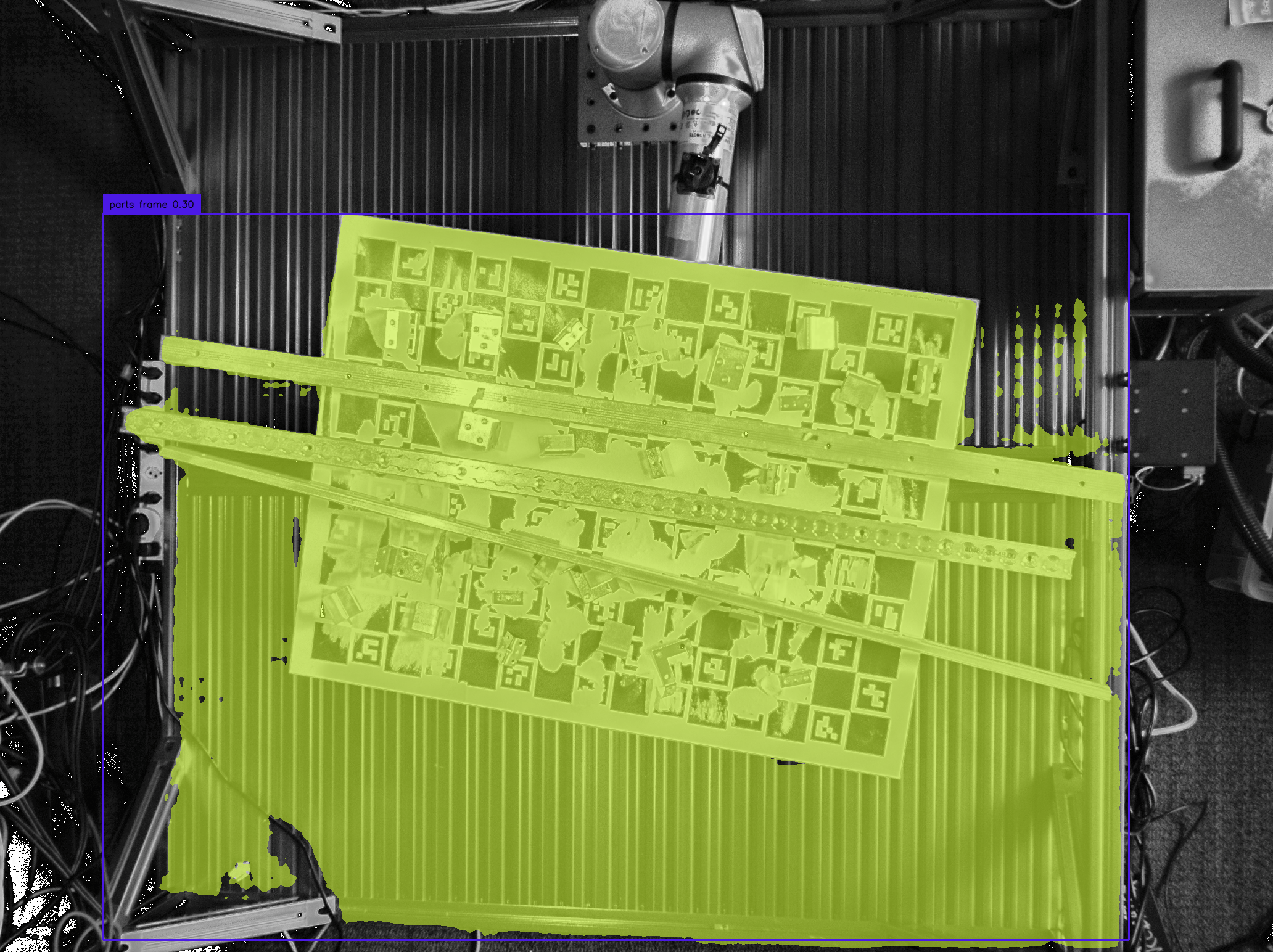}
  \end{subfigure}
  \begin{subfigure}[b]{0.32\linewidth}
    \includegraphics[width=\linewidth]{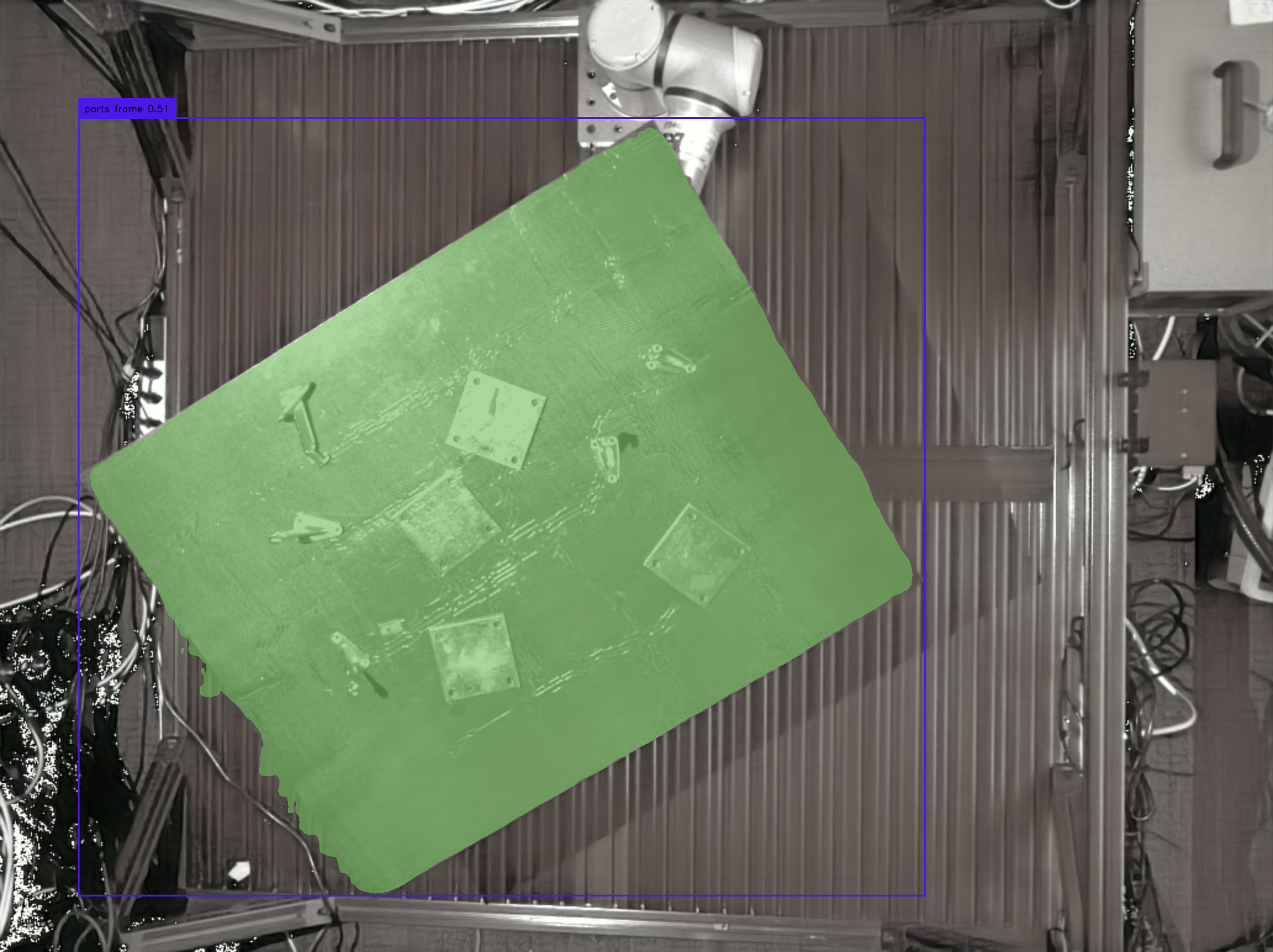}
  \end{subfigure}
\caption{Ablation study of the low-light enhancement model (LLEM). Row 1 shows GDINO without LLEM, and row 2 shows GDINO with LLEM.}
  \label{fig:ablation-llem}
\end{figure}

\begin{figure*}[htpb]
  \centering
  
  \begin{minipage}{0.03\linewidth}
    \hspace{\linewidth} 
  \end{minipage}\hfill
  \begin{minipage}[b]{0.15\linewidth}
    \centering
    {\footnotesize\textbf{Raw SAM output}}
  \end{minipage}\hfill
  \begin{minipage}[b]{0.15\linewidth}
    \centering
    {\footnotesize\textbf{GDINO ROI output}}
  \end{minipage}\hfill
  \begin{minipage}[b]{0.15\linewidth}
    \centering
    {\footnotesize\textbf{Raw SAM output}}
  \end{minipage}\hfill
  \begin{minipage}[b]{0.15\linewidth}
    \centering
    {\footnotesize\textbf{GDINO ROI output}}
  \end{minipage}\hfill
  \begin{minipage}[b]{0.15\linewidth}
    \centering
    {\footnotesize\textbf{Raw SAM output}}
  \end{minipage}\hfill
  \begin{minipage}[b]{0.15\linewidth}
    \centering
    {\footnotesize\textbf{GDINO ROI output}}
  \end{minipage}

  \vspace{1mm} 

  \begin{minipage}[c]{0.03\linewidth}
    \rotatebox{90}{{\footnotesize\textbf{IPD}}}
  \end{minipage}\hfill
  \begin{subfigure}[c]{0.15\linewidth}
    \includegraphics[width=\linewidth]{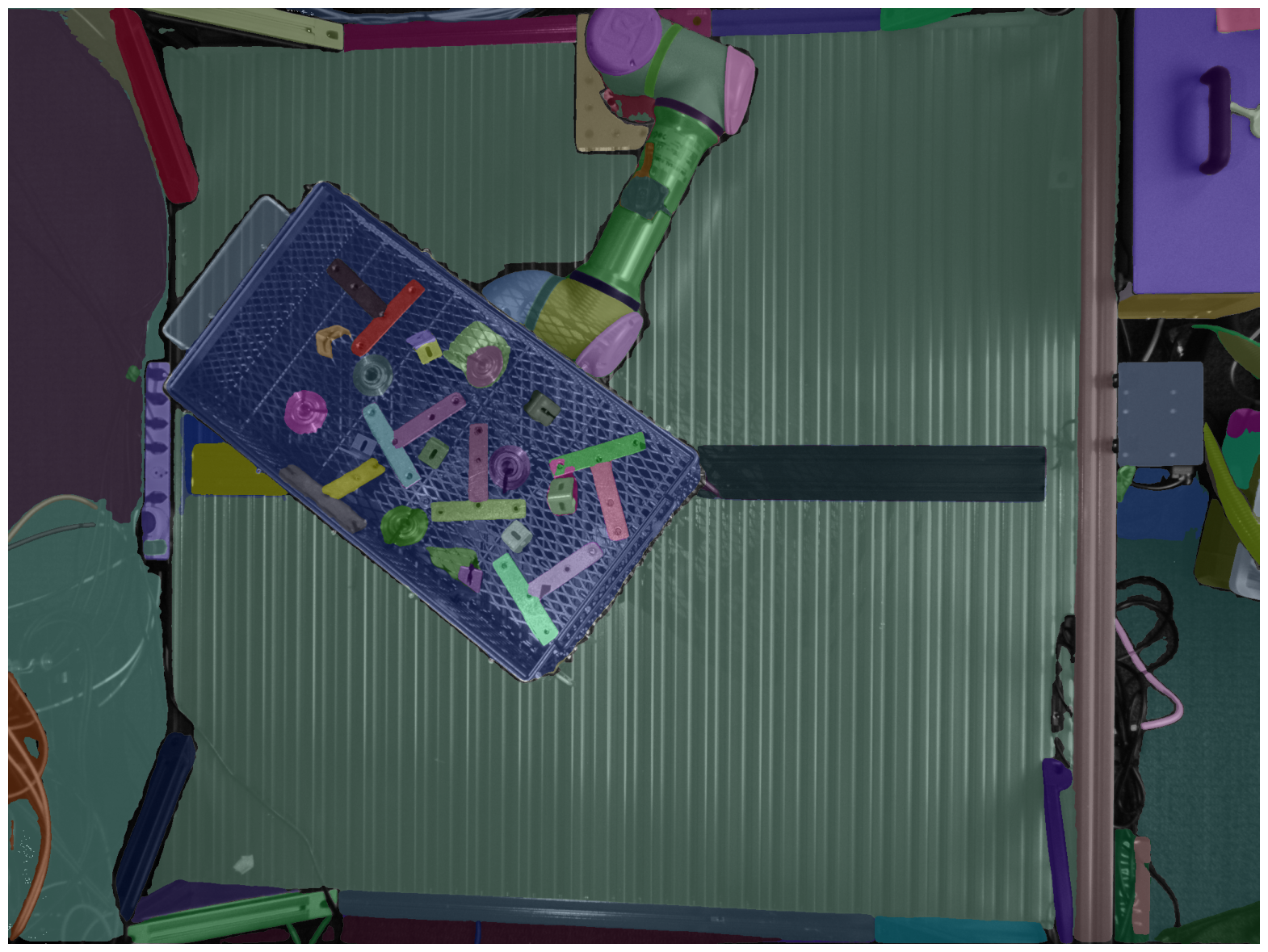}
  \end{subfigure}\hfill
  \begin{subfigure}[c]{0.15\linewidth}
    \includegraphics[width=\linewidth]{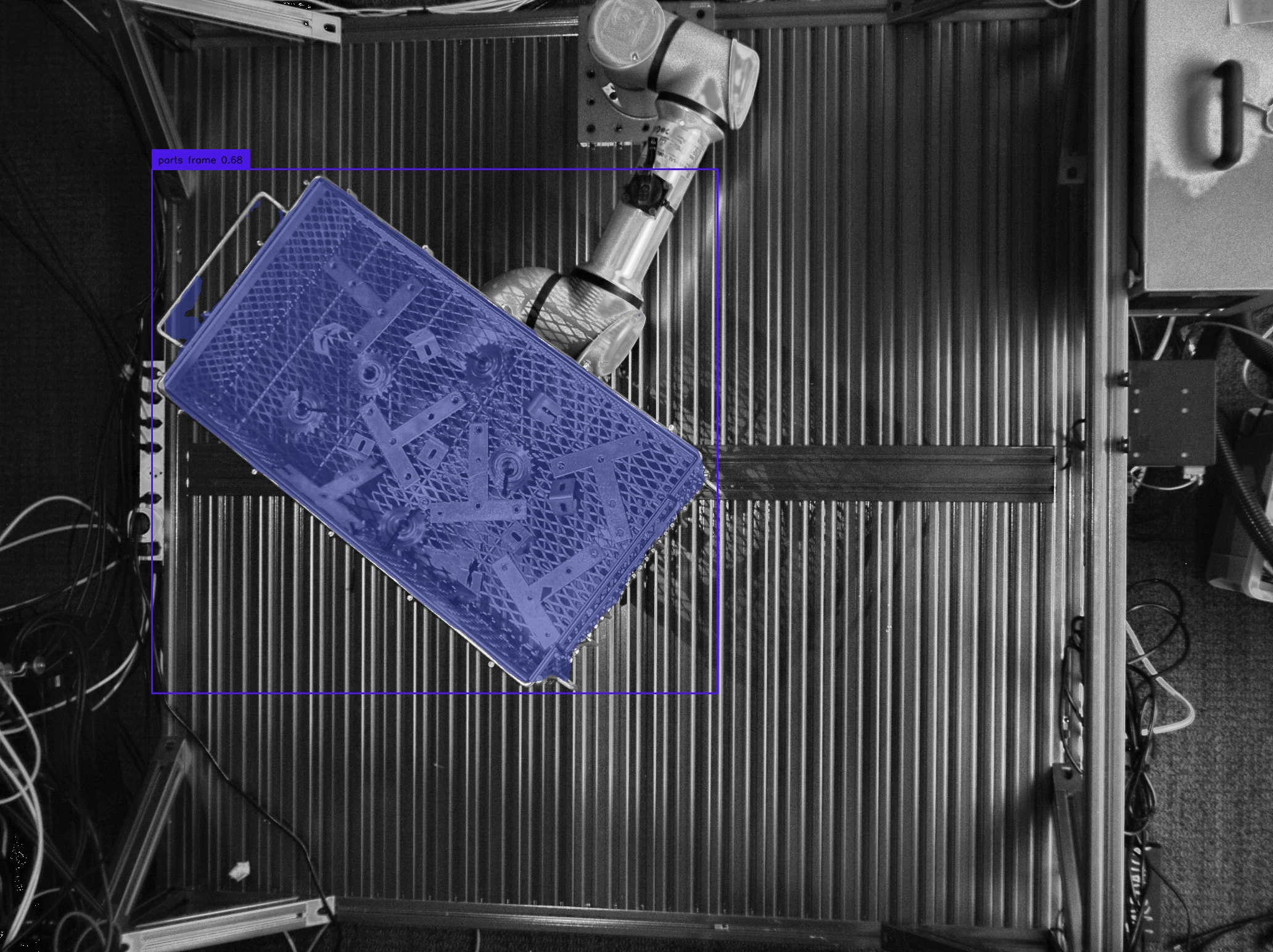}
  \end{subfigure}\hfill
  \begin{subfigure}[c]{0.15\linewidth}
    \includegraphics[width=\linewidth]{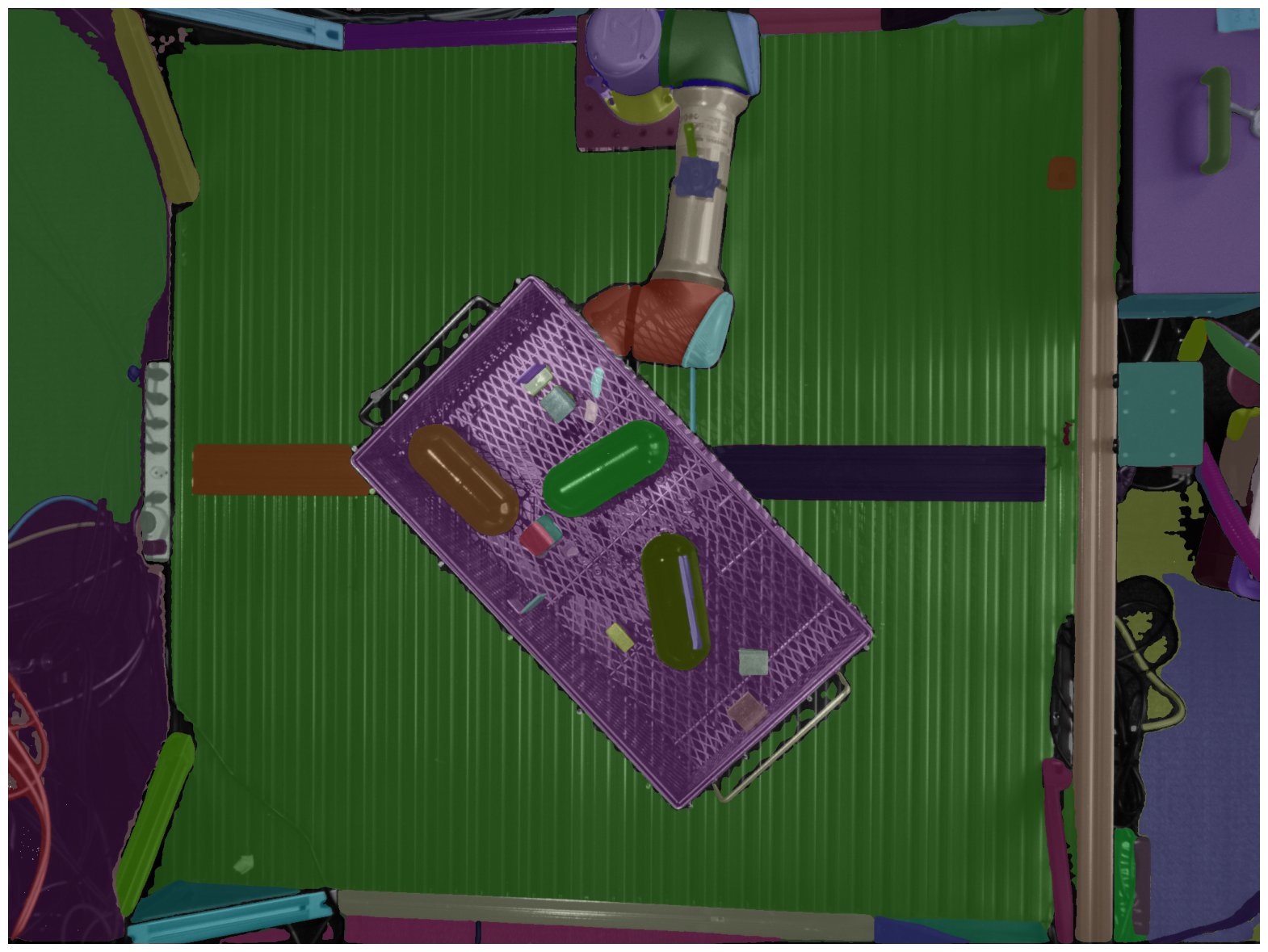}
  \end{subfigure}\hfill
  \begin{subfigure}[c]{0.15\linewidth}
    \includegraphics[width=\linewidth]{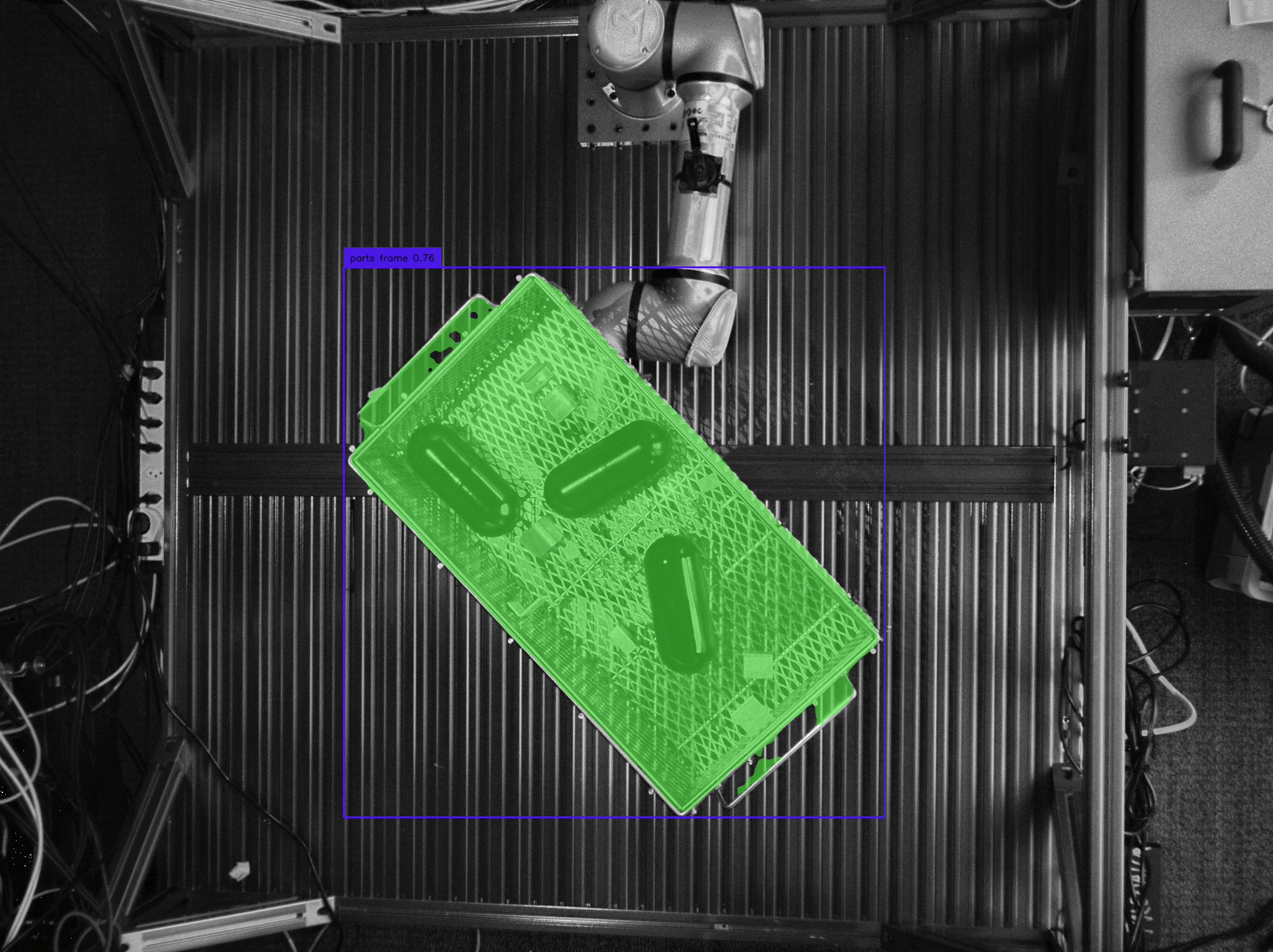}
  \end{subfigure}\hfill
  \begin{subfigure}[c]{0.15\linewidth}
    \includegraphics[width=\linewidth]{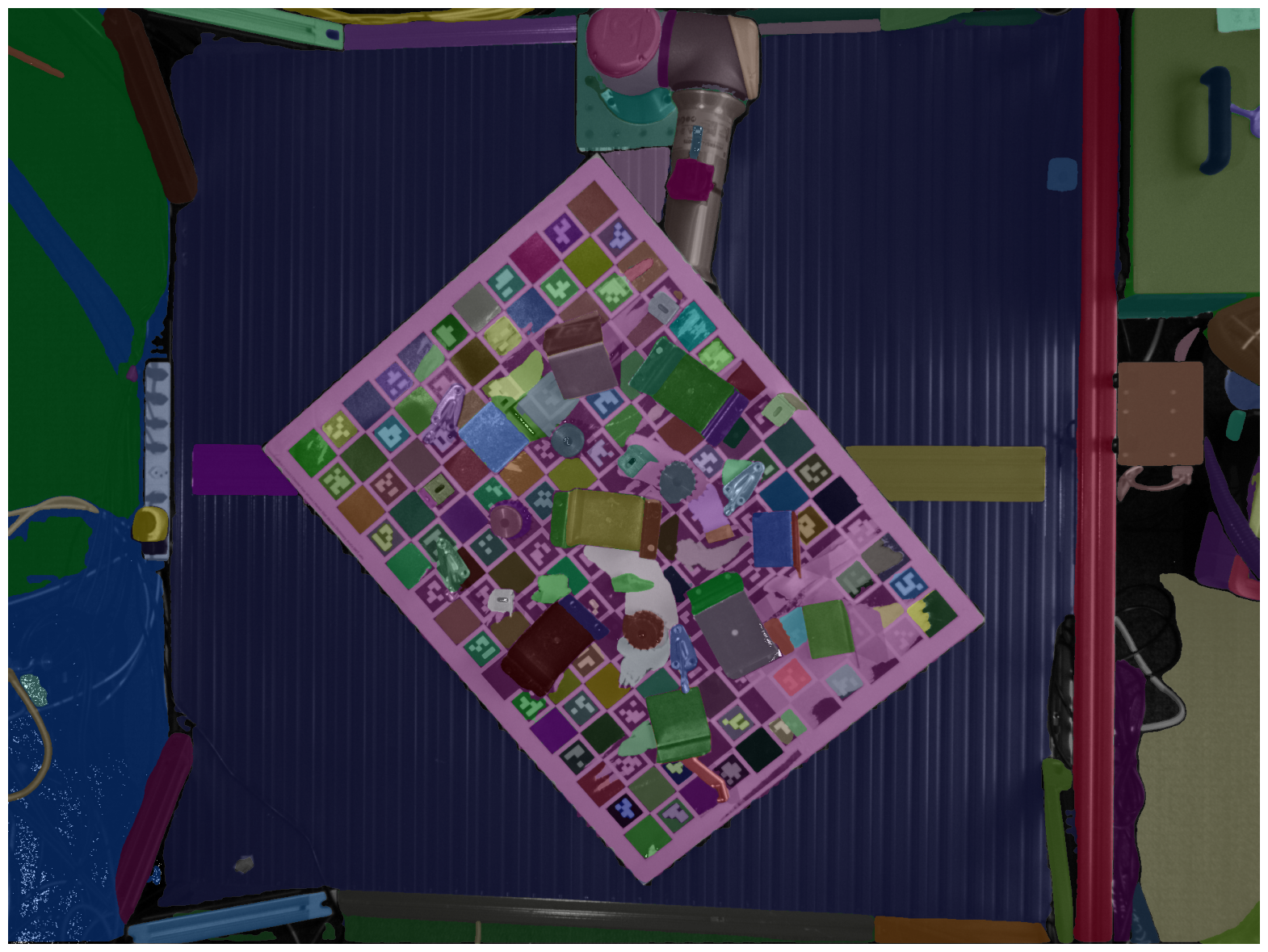}
  \end{subfigure}\hfill
  \begin{subfigure}[c]{0.15\linewidth}
    \includegraphics[width=\linewidth]{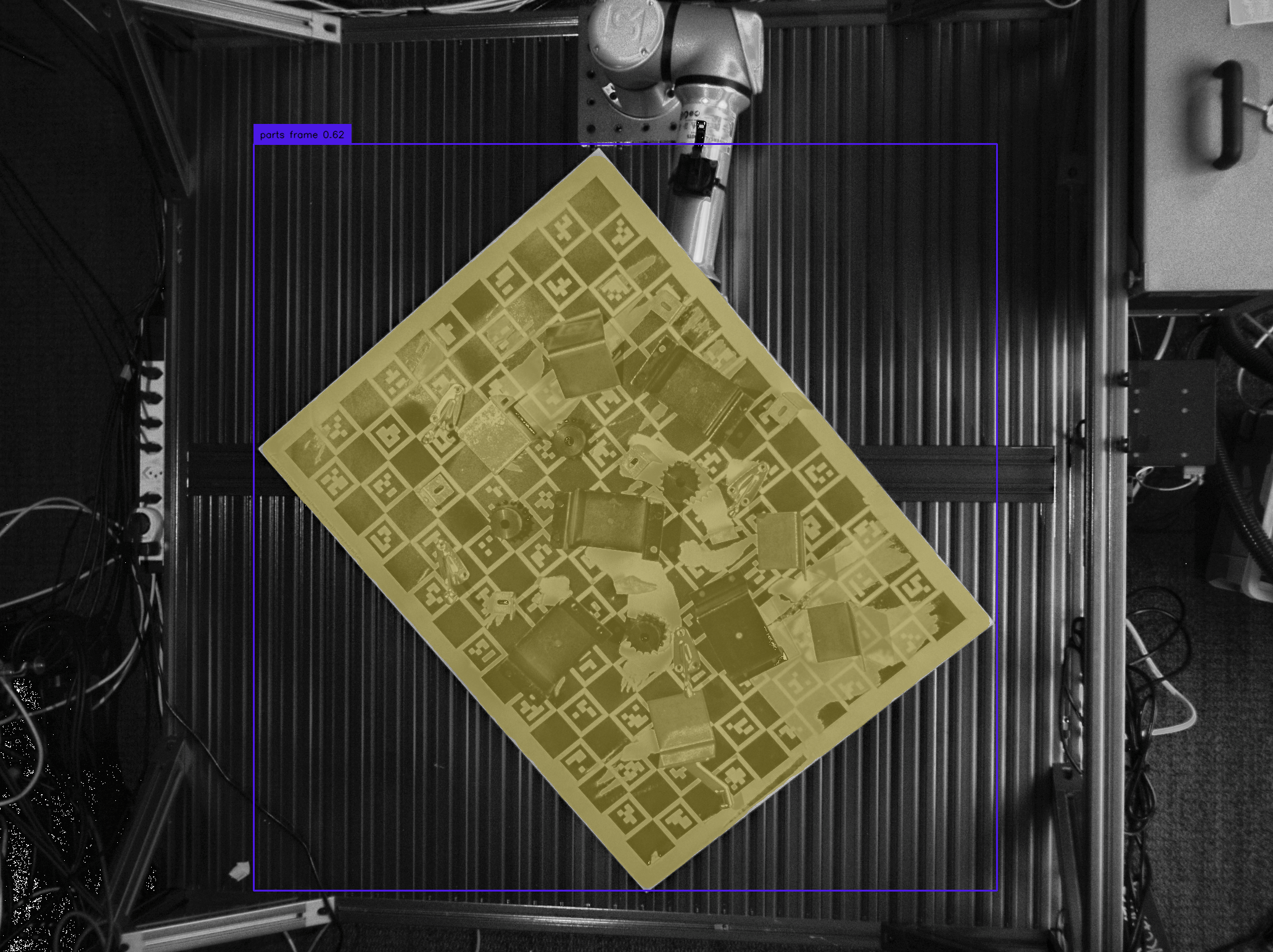}
  \end{subfigure}

  \vspace{2mm} 

  \begin{minipage}[c]{0.03\linewidth}
    \rotatebox{90}{{\footnotesize\textbf{XYZ-IBD}}}
  \end{minipage}\hfill
  \begin{subfigure}[c]{0.15\linewidth}
    \includegraphics[width=\linewidth]{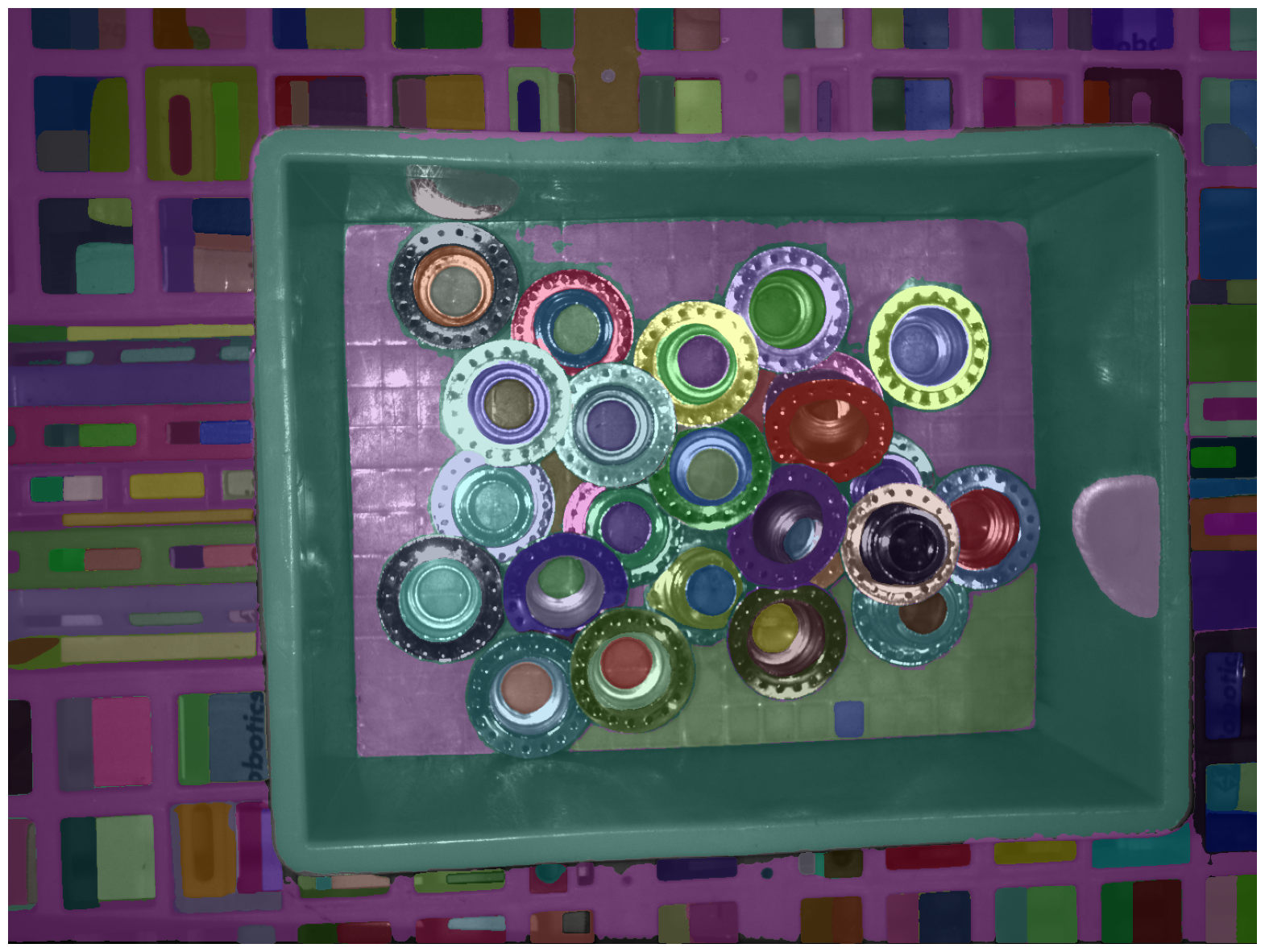}
  \end{subfigure}\hfill
  \begin{subfigure}[c]{0.15\linewidth}
    \includegraphics[width=\linewidth]{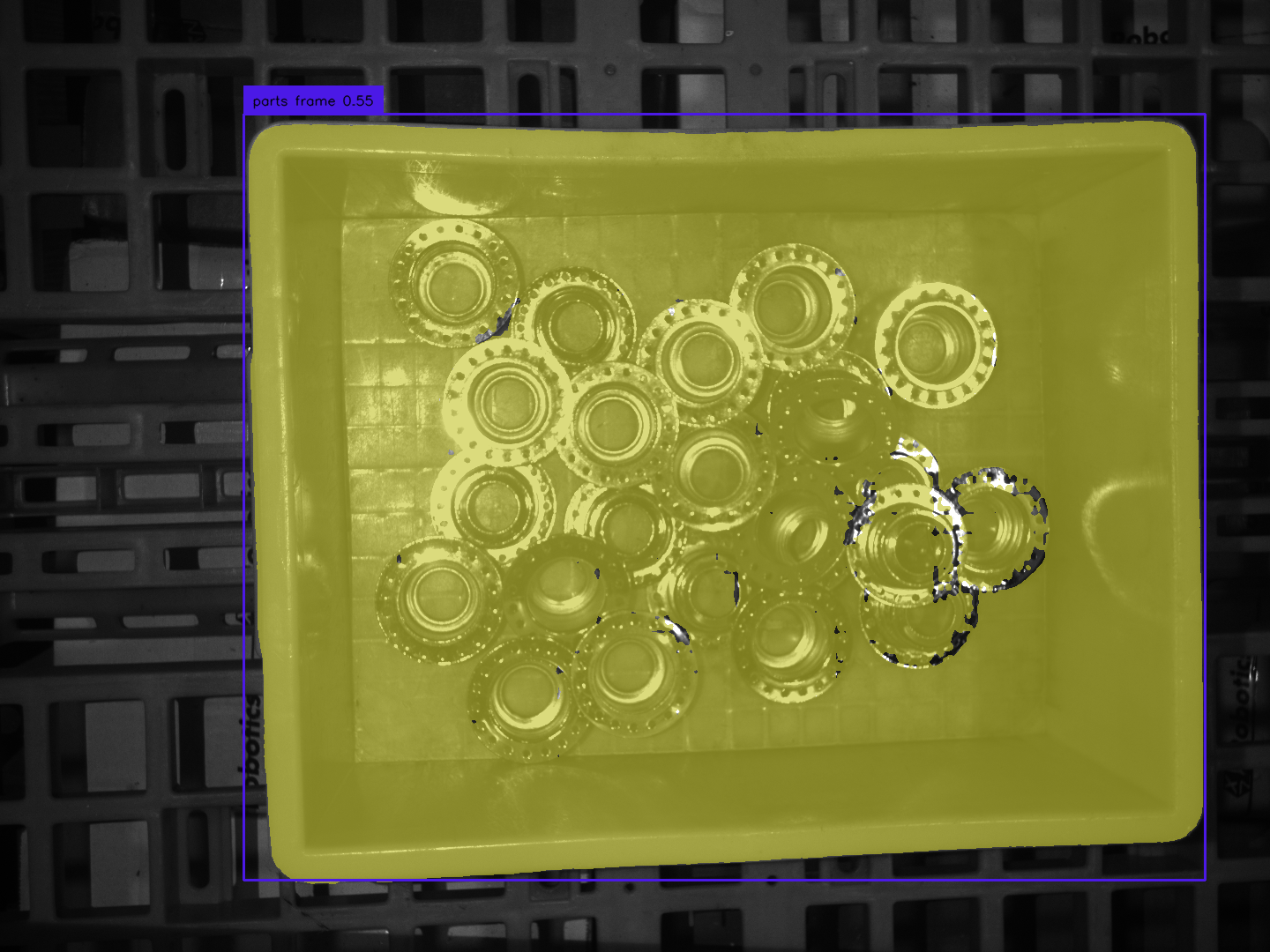}
  \end{subfigure}\hfill
  \begin{subfigure}[c]{0.15\linewidth}
    \includegraphics[width=\linewidth]{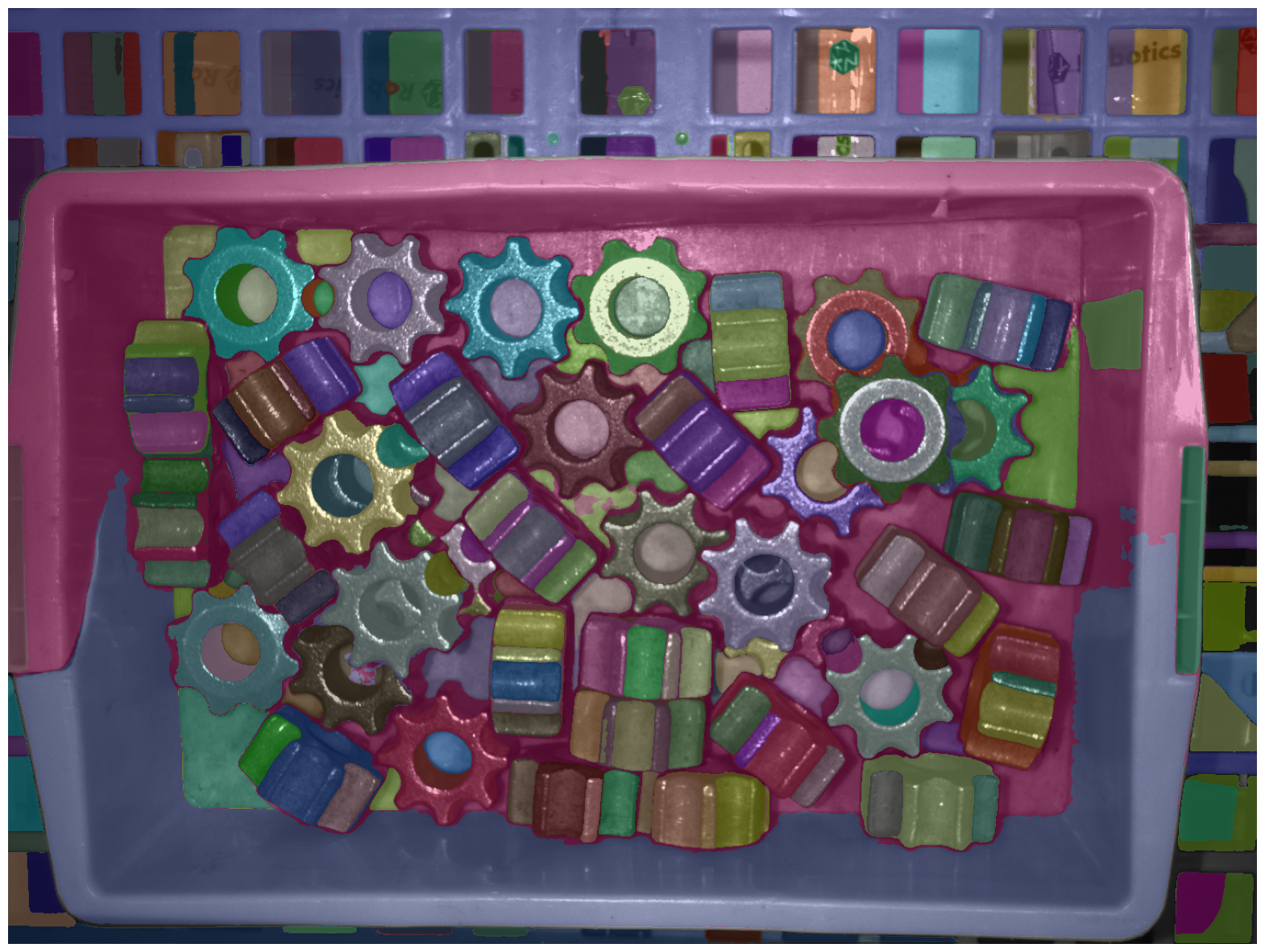}
  \end{subfigure}\hfill
  \begin{subfigure}[c]{0.15\linewidth}
    \includegraphics[width=\linewidth]{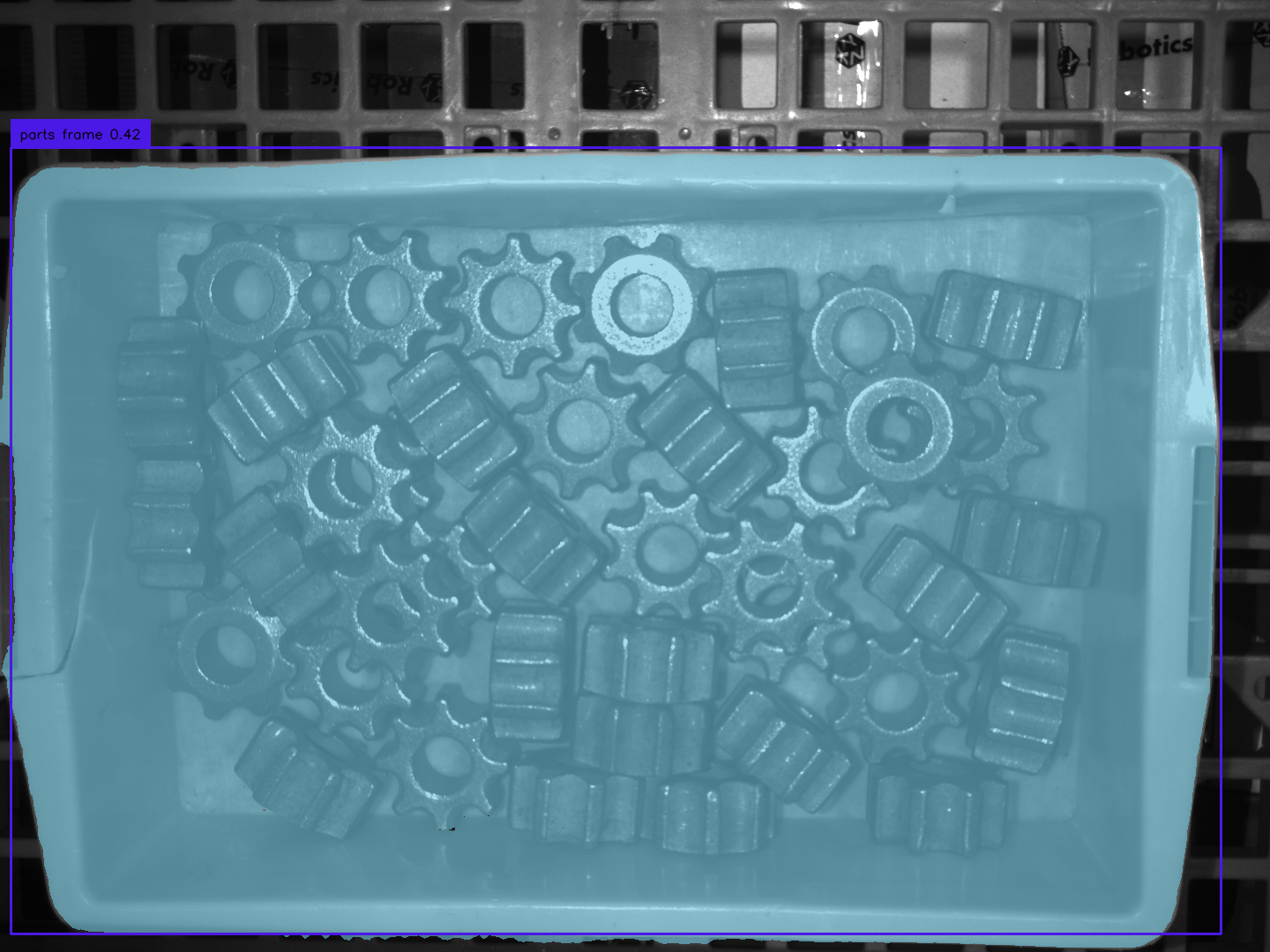}
  \end{subfigure}\hfill
  \begin{subfigure}[c]{0.15\linewidth}
    \includegraphics[width=\linewidth]{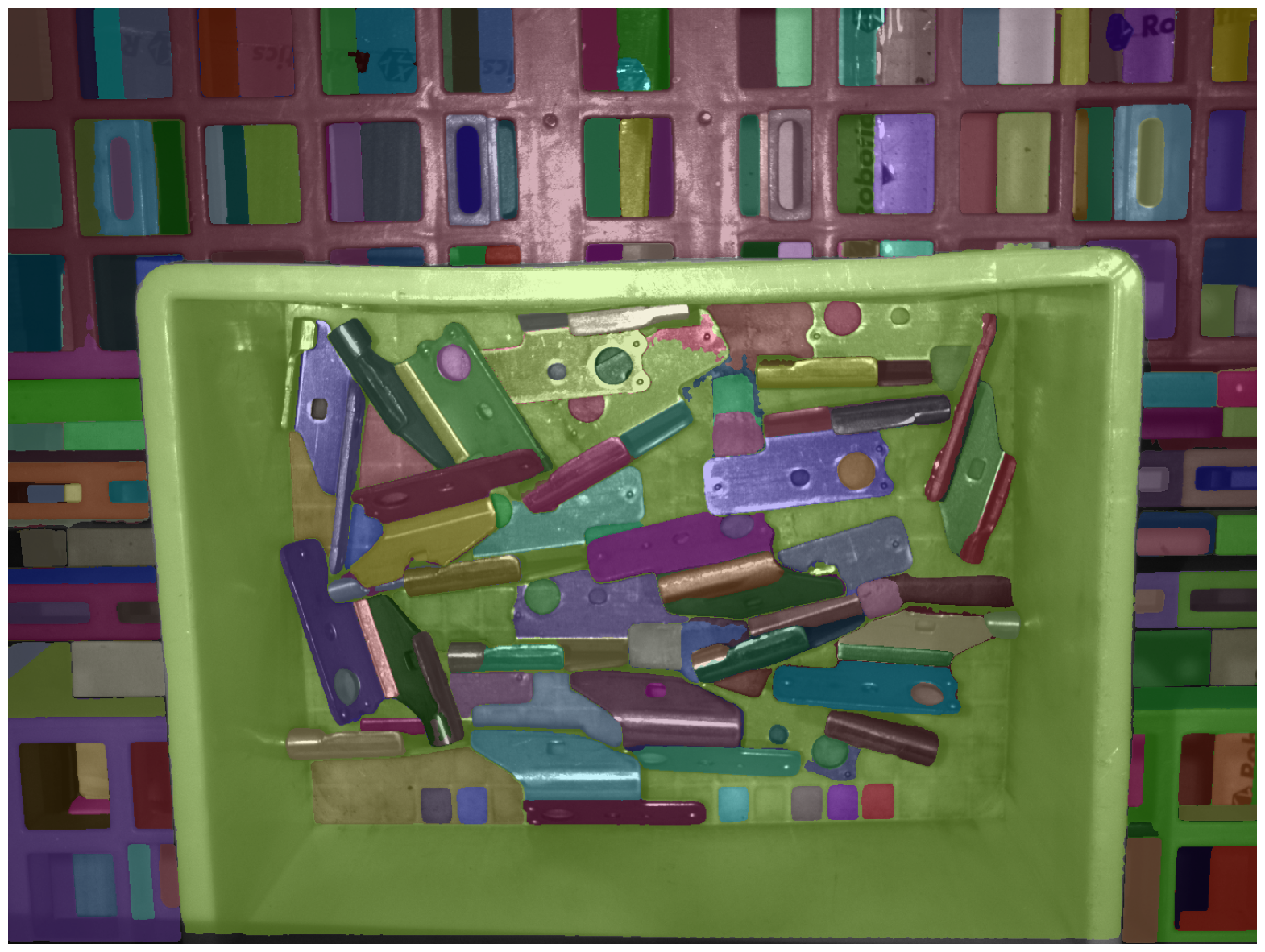}
  \end{subfigure}\hfill
  \begin{subfigure}[c]{0.15\linewidth}
    \includegraphics[width=\linewidth]{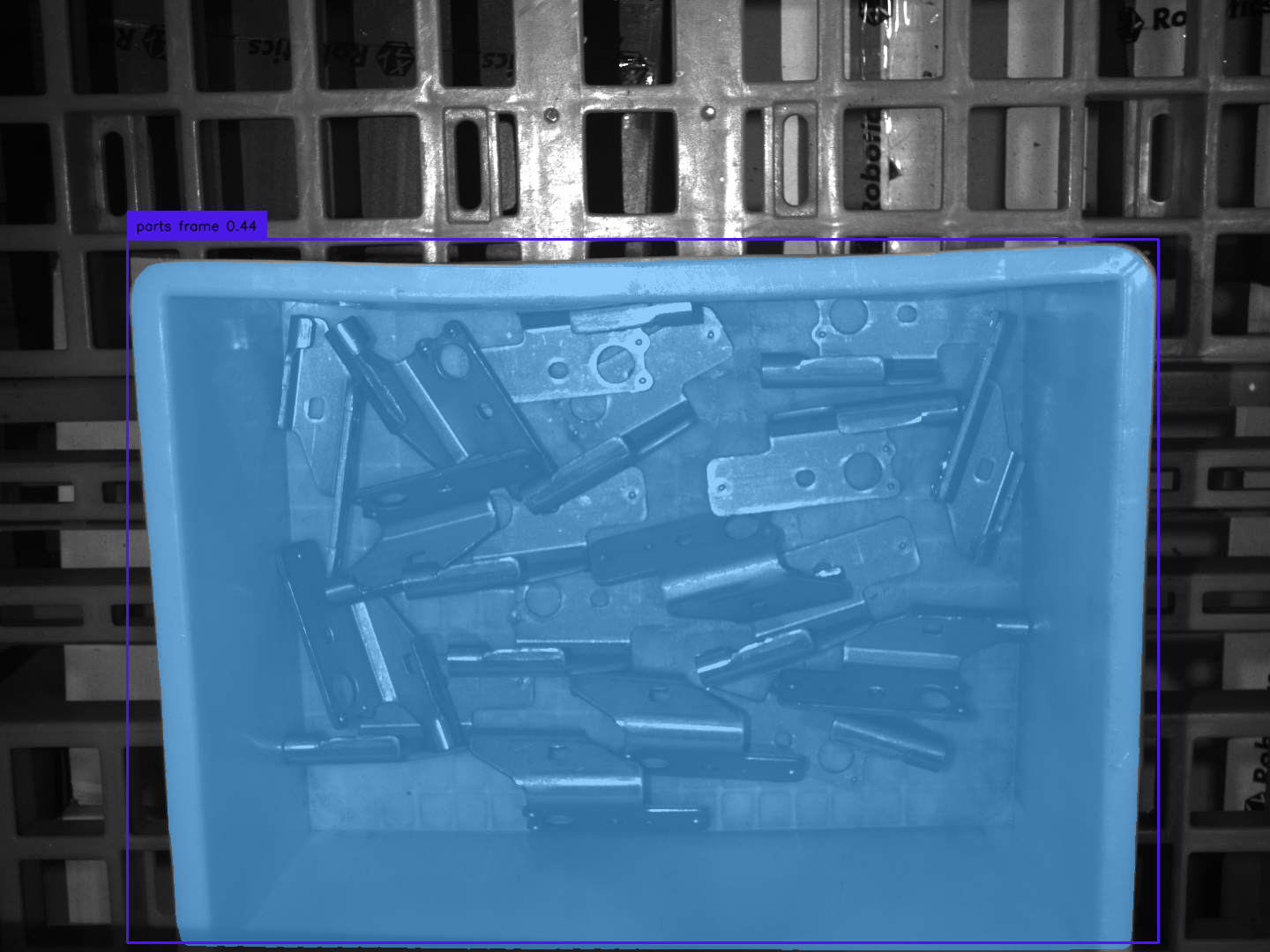}
  \end{subfigure}
  
  \caption{Qualitative comparison of raw SAM segmentation and GDINO-based ROI filtering. Top: results on the IPD dataset~\cite{Kalra_2024_CVPR}. Bottom: results on XYZ-IBD~\cite{huang2025xyz}. GDINO effectively removes background clutter using text prompts, producing cleaner regions of interest compared to raw SAM outputs.}
  \label{fig:sam_VS_gdino}
\end{figure*}

\subsection{Dataset}
We evaluate our detection pipeline on three BOP-Industrial benchmarks: XYZ-IBD~\cite{huang2025xyz}, IPD~\cite{Kalra_2024_CVPR}, and ITODD-MV~\cite{Drost_2017_ICCV}. ITODD-MV extends the ITODD dataset with multi-view calibration for the scenes, including 28 industrial parts under challenging geometry and lighting. IPD provides multi-modal captures from 13 calibrated cameras, including polarization and structured-light sensors, offering diverse lighting conditions and precise annotations. XYZ-IBD focuses on 15 metallic parts in bin-picking setups with heavy occlusion, random stacking, and clutter, making it particularly demanding for detection. Those three datasets are specifically designed for the industrial setup, with high-precision annotation and challenging conditions compared to the previous household scenarios. 

\subsection{Evaluation metric} We evaluate our approach using the metrics defined in the BOP Challenge \cite{nguyen2025bop,sundermeyer2023bopchallenge}. For the detection task, we report the Average Precision (AP), computed as the mean over Intersection-over-Union (IoU) thresholds from 0.50 to 0.95 in increments of 0.05. For the pose estimation task, we report the Average Recall (AR) with respect to three distinct error functions: Visible Surface Discrepancy (VSD), Maximum Symmetry-Aware Surface Distance (MSSD), and Maximum Symmetry-Aware Projection Distance (MSPD). 

\subsection{Implementation Details} We employed the HVI Model \cite{yan2025hvi} for low light enhancement by setting an intensity threshold of 50 so that only images with pixel values below this limit are processed, thereby avoiding overexposure. For background removal, we applied GDINO \cite{liu2023grounding} with the prompt “Parts frame where multiple parts inside it,” selected the crop with the highest confidence score, and discarded the rest. The resulting cropped image was then forwarded to the segmentation stage.

Following CNOS \cite{nguyen2023cnos} and SAM6D \cite{lin2023sam6d}, we employ the default ViT-H SAM \cite{kirillov2023segany} and FastSAM \cite{zhao2023fastsam} for mask proposal generation, and use the default ViT-L \cite{Dosovitskiy2020image} backbone of DINOv2 \cite{oquab2023dinov2} to extract class and patch embeddings.

\subsection{2D Detection of Unseen Objects }
In Table \redref{tab:results}, we demonstrate that our image preprocessing strategies outperform the baseline SAM6D \cite{lin2023sam6d} on the BOP Industrial dataset by averages of {\color{red}5.5\%} with FastSAM and {\color{red}6.2\%} with SAM. Specifically, the ITODD dataset features good lighting and no background clutter; therefore, we use the baseline model without any preprocessing. In contrast, the IPD dataset presents challenging lighting and a cluttered environment, where our preprocessing yields substantial gains of {\color{red}13.6\%} with FastSAM and {\color{red}17.3\%} with SAM.

\begin{table}[t]
  \centering
  \resizebox{\linewidth}{!}{%
  \begin{tabular}{llll}
    \toprule
    Method & Dataset & {Detection AP $\uparrow$} & {Pose AR $\uparrow$} \\
    \midrule
    SAM6D (SAM) & XYZ-IBD & 25.7 & 55.0 \\
    Ours (SAM)  & XYZ-IBD & 27.0 \textcolor{green}{(+1.3)} & 56.2 \textcolor{green}{(+1.2)} \\
    SAM6D (SAM) & IPD     & 30.0 & 40.9 \\
    Ours (SAM)  & IPD     & 47.3 \textcolor{green}{(+17.3)} & 61.9 \textcolor{green}{(+21.0)} \\
    \bottomrule
  \end{tabular}%
  }
  \caption{Comparison of pose estimation results using SAM6D segmentation and our segmentation as the detection input for the SAM6D pose estimation model.}
  \label{tab:ablation}
\end{table}

\begin{table}[t]
  \centering
  \begin{tabular}{lll}
    \toprule
    Input Modality & Segmentation Model & \textbf{AP $\uparrow$} \\
    \midrule
    Depth & FastSAM & 37.4 \\
    RGB & FastSAM & 42.2 \textcolor{green}{(+4.8)} \\
    Depth & SAM     & 42.6 \\
    RGB & SAM & 47.3 \textcolor{green}{(+4.7)} \\
    \bottomrule
  \end{tabular}
   \caption{We compare the performance of the Gounding-DINOdifferent on the IPD dataset, with different configurations on input modalities and SAM models.}
  \label{tab:ablation}
\end{table}

\subsection{Runtime Analysis}
In Table~\redref{tab:run_time}, we show the average runtime of our method compared with SAM6D~\cite{lin2023sam6d} on a single RTX4090 GPU, measured per image. We also report the runtime for each stage separately, including the preprocessing stage and the main detection stage. Although our method introduces an additional preprocessing stage compared to SAM6D, this stage does not increase the overall runtime much. Rather, it speeds up the main detection stage by processing cropped images of lower resolution.
\begin{table*}[!t]
\centering
\small
\setlength{\tabcolsep}{4pt}
\begin{tabular}{llccc}
\toprule
\multicolumn{2}{l}{} & \multicolumn{3}{c}{\textbf{Runtime (s) $\downarrow$}}\\
\cmidrule(lr){3-5}
\textbf{Dataset} & \textbf{Method} &
\multicolumn{1}{c}{\textbf{Preprocessing}} &
\multicolumn{1}{c}{\textbf{Proposal + Matching}} &
\multicolumn{1}{c}{\textbf{Total}} \\
\midrule
\makecell[l]{\textbf{XYZ-IBD} (1440×1080)} & SAM6D (FastSAM) & {N/A} & 0.479 & 0.479 \\
& SAM6D (SAM)     & {N/A} & 2.367 & 2.367 \\
& Ours (FastSAM)  & 0.176 & 0.428 (\textcolor{green}{-0.051}) & 0.604 (\textcolor{red}{+0.125}) \\
& Ours (SAM)      & 0.139 & 1.943 (\textcolor{green}{-0.424}) & 2.082 (\textcolor{green}{-0.285}) \\
\midrule
\makecell[l]{\textbf{IPD} (2064×1544)} & SAM6D (FastSAM) & {N/A} & 0.405 & 0.405 \\
& SAM6D (SAM)     & {N/A} & 2.062 & 2.062 \\
& Ours (FastSAM)  & 0.145 & 0.351 (\textcolor{green}{-0.054}) & 0.496 (\textcolor{red}{+0.091})\\
& Ours (SAM)      & 0.120 & 1.955 (\textcolor{green}{-0.107}) & 2.075 (\textcolor{red}{+0.013}) \\
\bottomrule
\end{tabular}
\caption{\textbf{Run-time.} We report the run-time of the preprocessing stage and the main detection stage (proposal generation and matching) on a single RTX4090 GPU. We also compare our method with SAM6D \cite{lin2023sam6d} about the run time on dataset XYZ-IBD \cite{huang2025xyz} and IPD \cite{Kalra_2024_CVPR}.}
\label{tab:run_time}
\end{table*}

\subsection{6D Detection of Unseen Objects }
Table \redref{tab:ablation} shows the effect of our preprocessing-based segmentation on subsequent 6D pose estimation. On the XYZ-IBD dataset, our method provides improvements over SAM6D (Detection AP: {\color{red}+1.3\%}, Pose AR: {\color{red}+1.2\%}). On the IPD dataset, characterized by extreme HDR lighting and background clutter, our approach yields substantial gains of {\color{red}+17.3\%} Detection AP and {\color{red}+21.0\%} Pose AR. These results confirm that enhancing segmentation quality through preprocessing directly translates into more reliable pose estimation, especially in complex industrial scenes.

\subsection{Ablation study}
We conduct ablation studies on the IPD \cite{Kalra_2024_CVPR} and XYZ-IBD \cite{huang2025xyz} datasets to evaluate the efficacy of individual strategy in image preprocessing with the mask prediction generated by FastSAM. 

\textbf{Efficacy of LLEM.} Although using only the low-light enhancement module yields virtually no performance gain, we find that it aids GDINO in achieving more accurate background removal, as illustrated in Figure \redref{fig:ablation-llem}.

\textbf{Efficacy of GDINO.} Using GDINO, we can effectively remove the chaotic background, preventing it from being detected as a target object. As shown in Table \redref{tab:results}, using only the GDINO module outperforms the baseline by 13.1\% on IPD and 2.9\% on XYZ-IBD, respectively. Figure \redref{fig:sam_VS_gdino} shows the qualitative results of GDINO on datasets IPD and XYZ-IBD with both depth and RGB input modality.

\textbf{GDINO input modality.} GDINO supports either RGB or depth modalities. Optionally, when accurate depth measurements are available, we can exploit geometric priors for background filtering. For the IPD dataset \cite{Kalra_2024_CVPR}, we preprocess the depth information by first masking any pixels outside the specified working distance of 150 cm to 200 cm. The remaining depth values are then converted into a plasma colormap to enhance layer separation. Grounding-DINO then processes this pseudo-color depth map with the same prompt, generating a bounding box that refines ROI localization.
Table \redref{tab:ablation} demonstrates that using RGB yields an AP improvement of 4.3\% with FastSAM and 4.4\% with SAM compared to depth input. Specifically, we found that the depth map in the IPD dataset is less accurate, leading to the misclassification of target regions as background.

In Table~\ref{tab:run_time}, we show the average runtime of our method compared with SAM6D~\cite{lin2023sam6d} on a single RTX4090 GPU, measured per image. We also report the runtime for each stage separately, including the preprocessing stage and the main detection stage. Although our method introduces an additional preprocessing stage compared to SAM6D, this stage does not increase the overall runtime much. Rather, it speeds up the main detection stage by processing cropped images of lower resolution.
\section{Conclusion}
Object detection has become a bottleneck for the pose estimation methods under complex scenes. In this paper, we propose a simple yet effective 2D detection pipeline for unseen objects in challenging industrial scenarios. Based on the current generalizable detection baselines, we boost the performance by reducing the domain gap between image segments and rendered templates, and suppressing redundant background segments to minimize false candidates for object matching. Our approach combines a low‑light enhancement model with a background removal module to further boost the recent state-of-the-art method for a large margin under challenging lighting and cluttered environments. Our pipeline is applicable to any existing baselines, and the evaluation on the BOP-Industrial datasets demonstrates the effectiveness of the proposed method. 
{
    \small
    \bibliographystyle{ieeenat_fullname}
    \bibliography{main}
}
\end{document}